%% file: OPT-GAN.tex
\documentclass[letterpaper]{article} 
\usepackage{aaai23}  

\pdfoutput=1

\usepackage{xcolor}
\usepackage{amsmath,amsfonts}

\usepackage{times}  
\usepackage{helvet}  
\usepackage{courier}  
\usepackage[hyphens]{url}  
\usepackage{graphicx} 
\usepackage{natbib}  
\usepackage{caption} 
\usepackage{algorithm}
\usepackage{algorithmic}
\usepackage{booktabs}

\usepackage{pdfpages}

\usepackage{newfloat}
\usepackage{listings}
\DeclareCaptionStyle{ruled}{labelfont=normalfont,labelsep=colon,strut=off} 
\floatstyle{ruled}
\newfloat{listing}{tb}{lst}{}
\floatname{listing}{Listing}

\title{OPT-GAN: A Broad-Spectrum Global Optimizer for Black-box Problems by Learning Distribution}
\author{
    Minfang Lu \textsuperscript{\rm 1,5,}\equalcontrib,
    Shuai Ning \textsuperscript{\rm 1,2,}\equalcontrib,
    Shuangrong Liu \textsuperscript{\rm 3},
    Fengyang Sun \textsuperscript{\rm 4},
    Bo Zhang \textsuperscript{\rm 1,2}, \\
    Bo Yang \textsuperscript{\rm 2},
    Lin Wang \textsuperscript{\rm 1,}\footnote{Corresponding author: Lin Wang (wangplanet@gmail.com)}
}
\affiliations{
    \textsuperscript{\rm 1} Shandong Provincial Key Laboratory of Network Based Intelligent Computing,\\ University of Jinan, Jinan 250022, China \\
    \textsuperscript{\rm 2} Quan Cheng Laboratory, Jinan 250100, China \\
    \textsuperscript{\rm 3} Department of Computer Science, The University of Suwon, Hwaseong 18323, South Korea \\
    \textsuperscript{\rm 4} Victoria University of Wellington, Wellington 6140, New Zealand \\
    \textsuperscript{\rm 5} Cainiao Network, Hangzhou, China \\
}

\usepackage{bibentry}

\begin{document}

\maketitle
\begin{abstract}
{
Black-box optimization (BBO) algorithms are concerned with finding the best solutions for problems with missing analytical details. Most classical methods for such problems are based on strong and fixed \emph{a priori} assumptions, such as Gaussianity. However, the complex real-world problems, especially when the global optimum is desired, could be very far from the \emph{a priori} assumptions because of their diversities, causing unexpected obstacles.   In this study, we propose a generative adversarial net-based broad-spectrum global optimizer (OPT-GAN) which estimates the distribution of optimum gradually, with strategies to balance exploration-exploitation trade-off. It has potential to better adapt to the regularity and structure of diversified landscapes than other methods with fixed prior, e.g., Gaussian assumption or separability. 
Experiments on diverse BBO benchmarks and high dimensional real world applications exhibit that OPT-GAN outperforms other traditional and neural net-based BBO algorithms. The code and Appendix are available at {https://github.com/NBICLAB/OPT-GAN}
}

\end{abstract}
  
\section{Introduction}
\label{sect-Intro}

Optimization is a study of finding the best solutions to a given problem. In this field, global optimization searches for the globally best solution among all possible ones. We are interested in BBO problems:
  {
  \begin{equation}
      {\mathop{\arg \min }} \, f(\boldsymbol{x}):\Omega \to \mathbb{R}, \, \forall \boldsymbol{x}\in \Omega \subset \mathbb{R}^n,
  \end{equation}
  }%
where $n$ is the dimensionality of the searching domain $\Omega$. The ``black box'' means we have little information about $f$ but the function value $f(\boldsymbol{x})$ indicating the quality of the candidate solution $\boldsymbol{x}$. 

In terms of stochastic black-box optimization, we define $p(\boldsymbol{x})$ as a distribution representing the estimated chance that the global optimum $\boldsymbol{x}^*$ can be found at $\boldsymbol{x}$. Thus, the optimization can be viewed as progressively reshaping explicit/implicit  $p(\boldsymbol{x})$ based on sampled $\boldsymbol{x}$ and corresponding $f(\boldsymbol{x})$, toward the Dirac delta distribution $\delta _{x^ * }(\boldsymbol{x})$ with a spike at $\boldsymbol{x}^*$. Note that different optimizer reshapes $p(\boldsymbol{x})$ in different way (see Appendix A for the general framework).

The No Free Lunch Theorem (NFLT) \cite{Wolpert1997NFL} indicates the significance of prior knowledge to a global optimizer, as the performance difference between optimizers depends on how their \emph{a priori} assumptions match with the inherent regularities of problem. Although there are some specialized \emph{narrow-spectrum}\footnote{Inspired by the term broad-spectrum antibiotics, spectrum implies the range of applicable problems of an optimizer. The broad-spectrum ones expect to achieve acceptable performance at a wide range via strong configurability. However, a narrow-spectrum one aspires to the best performance at specific families of problems.} optimizers designed for specific problems by utilizing prior knowledge, e.g., Iterative Closest Point algorithm \cite{Chetverikov2002ICP} focuses on point-set matching related problems, the difficulty for obtaining sufficient problem-related information in real world leads to prosperity of \emph{broad-spectrum} optimizers.

\input{fig_tex/fig_sketch.tex}

To deal with the challenge of information deficiency, the broad-spectrum optimizers maximize performance with little \emph{a priori} knowledge on target black-box optimization problems. They expect to achieve acceptable performance at various problems via strong configurability, i.e., balancing exploration-exploitation (E-E) trade-off using configurable parameters or modules.
  
There are two branches of broad-spectrum optimizers: \emph{model-building} and \emph{model-free} algorithm, e.g., {Genetic Algorithms (GA)}  \cite{Holland1975GA} and {Particle Swarm Optimization (PSO)} \cite{Eberhart1995PSO}, directly manipulate solutions in an implicit way. Despite wide application, they are weak in capturing complex problem structures \cite{katoch2021review, liu2017multi}. The model-building ones search with explicit models, such as modeling landscape (i.e., $f$) or gradient (i.e., $\nabla f$). Among them, modeling $p(\boldsymbol{x})$ directly is one of primary branches, e.g., Estimation of Distribution Algorithms (EDAs) \cite{Hauschild2011EDAsurvey}, which have achieved considerable advances \cite{EDA_advance, NIPS1996_4c22bd44, chen2017personalized}. They have potentials to capture complex structures and learn accurate distribution of promising solutions without much attention on landscape details \cite{cheng2018model, antoniou2021differential}.

\setcounter{footnote}{0}
However, the diversity of real-world problems with massive unknown properties makes choosing proper \emph{a priori} assumption for EDAs difficult. For instance, as a widely used EDA method, Evolutionary Strategy with Covariance Matrix Adaptation (CMA-ES) fails to reshape its distribution type for problem whose landscape is distinct from a Gaussian surface or mixture of Gaussians \cite{Liu2020Bending}.

Without violating the NFLT, a “shapeshifter” broad-spectrum  optimizer with less dependency on prior of target problems is desired, especially a flexible $p(\boldsymbol{x})$ EDA method, which tunes distribution to adapt to diversified problems progressively.
As an universal distribution learner \cite{Lu2020UAT}, the Generative Adversarial Networks (GANs) \cite{Goodfellow2014GAN} is one of the strongest candidates for capturing arbitrary distribution. In fact, their effectiveness for optimization have been preliminarily verified by GAN-based Solver (GBS)
\cite{gillhofer2019gan} for \emph{local optimization} and MOE-GAN \cite{MOE-GAN} for \emph{multiobjective optimization}.

\subsubsection{Motivation} 
Unfortunately, the \emph{global optimization} by estimating the distribution of optima progressively is still hard to reach, as we have to carefully balance the exploration-exploitation trade-off during optimization to gradually expose global optimum. This problem is even more notable for black-box tasks because we have no information on which side is better.
  
A question arises here: \emph{can we design a broad-spectrum global optimizer  by modeling a flexible $p(\boldsymbol{x})$ estimator, enabling balancing E-E trade-off and adapting to arbitrary black-box problem?}

\subsubsection{Contribution} We propose a generative adversarial net-based broad-spectrum optimizer for global optimization, named OPT-GAN, which estimates the distribution of optimum gradually by balancing E-E trade-off (see Fig. \ref{fig:schematic}). OPT-GAN shows potential to better adapt to the structures of different landscapes than other EDA methods with fixed prior, e.g., Gaussian distribution assumption in CMA-ES.
  Detailed contributions are listed as follows:
  \begin{itemize}
  \item A novel GAN-based broad-spectrum global optimizer for black-box problems is proposed to estimate the distribution of global optimum gradually.
  \item As a global optimizer, OPT-GAN adopts a bi-discriminators framework to guide the generator to learn balancing E-E trade-off during optimization.
  \item A continually updating and shrinking optimal set provides data about regularities, encouraging exploration at the early stage and exploitation at the late stage. 
  \item To avoid biased initialization and premature convergence, a generator pre-training method is used to ensure full domain initialization.
  \item Experiments manifest that OPT-GAN achieves the best results on diversified problems compared with different types of optimizers, including neural network-based state-of-the-art optimizers.
  \end{itemize}

  \section{Related Works}
  \label{sect-RelatedWorks}

  \subsubsection{Narrow-Spectrum Optimizers} Narrow-spectrum optimizers can solve specific problems in an 
  efficient way by exploiting the problem-related knowledge \cite{Serafino2014optimizingNFL}, e.g., Backpropagation \cite{Lecun1988BP}, 
  FBGAN \cite{FBGAN}, CbAs \cite{CBAS}, DbAs \cite{DBAS}, MetricGAN \cite{fu2019metricgan}, and PGATS \cite{PGATS}. However, 
  such prior knowledge or analytical details are often not available for many real-world optimization problems \cite{Dulacarnold2019ChallengesRL, thor2020generic}.

 \subsubsection{Model-Free Optimizers} As a typical family of broad-spectrum optimizers, model-free algorithms manipulate solutions directly. This family involves, e.g., BFGS \cite{Liu1989BFGS}, Nelder-Mead method (NM) \cite{Nelder1965Simplex}, Generalised Pattern Search \cite{Torczon1997GPS}, Simulated Annealing \cite{SimulatedAnnealing}, GA, and PSO. They are popular for their approachable and easily implementable mechanism, yet suffering from weak mathematical structure and low efficiency \cite{Audet2017DFOtextbook}. 
  
\input{fig_tex/fig_flowchart.tex}
  
 \subsubsection{Model-Building Optimizers} 
 Model-building optimizers rely on adjusting an explicit model to search the BBO problems.
 \textbf{(1) Surrogate Optimization Algorithms} simulate the landscape to reduce practical evaluations, e.g., Response Surface Methods \cite{Jones2001RSM} and Support Vector Machine-based Surrogate Models \cite{Ciccazzo2015SVMsurrogate}. Bayesian Optimization adopts Gaussian processing to accurately regress the landscape, and has been successfully applied to many expensive but low dimensional problems \cite{balandat2020botorch,shahriari2015taking,eriksson2019scalable}. However, when solving high-dimensional problems, sample size that accurate modeling requires increases exponentially, leading to degraded sampling and inefficient search \cite{raponi2020high, luo2012gaussian}. In addition, its time consumption increases drastically when reconstructing the model with increasing samples. Although some approaches such as incremental learning could alleviate this issue \cite{jenatton2017bayesian,klein2017fast}, extra flaw of catastrophic forgetting limits its learning ability. 
 textcolor{blue}{Note that the discussions are limited to navie Bayesian Optimization.} 
 \textbf{(2) Gradient Estimators}, such as OPEN-ES \cite{salimans2017evolution}, and Adaption Directional Gaussian Smoothing \cite{tran2020adadgs}, perform well on quasi-convex problems due to local learning on gradient but poorly on non-convex problems. 
  
\textbf{(3)} The last family \textbf{EDAs} estimate the distribution of optima by progressively sampling solutions, such as Cross Entropy Method \cite{Rubinstein2013CEM} and Latent Space-based EDA \cite{dong2019latent}. As one of the most prominent EDAs, CMA-ES estimates the better region by adaptively reshaping the Gaussian model and has received lots of successful stories \cite{Kampf2009hybrid,Loshchilov2016CMAESnn}. 
  EDAs have presented remarkable performance on the problems  compatible with their distribution assumption, e.g., Gaussian distribution. Nevertheless, EDAs with strong \emph{a priori} are weak in adapting to diversified real-world problems.   

  \subsubsection{Neural Network-based Optimizers} The neural network(NN)-based  black-box optimizers have recently gained growing attention due to their powerful approximation ability, e.g., RandomRL \cite{Mania2018randomRL}, GradientLess Descent \cite{Golovin2019gradientless}, and Indirect Gradient Learning (IGL) \cite{Lillicrap2019IGL}. Explicit Gradient Learning (EGL) \cite{Sarafian2020EGL} directly estimates the gradient $\nabla f$ by learning the parametric weights. In addition, Weighted Retraining (WR) \cite{NEURIPS2020_81e3225c} adopts latent manifold learning to convert a problem landscape into a latent space via generative models and search by traditional optimizer in the new space \cite{GNN_CMA_ES,GNN_ES,CMA_decoder}.

Given the universal distribution learning capability of GANs, the GAN-based optimizer may herald a fruitful direction by  modeling $p(\boldsymbol{x})$. In 2019, focusing on inverse problem, a GAN-based Solver \cite{gillhofer2019gan} was proposed. Although it's a preliminary work and only tested on topology optimization, the potential of GANs have been exhibited. If we temporally move our attention to another related field \emph{multiobjective optimization}, there is also a GAN-based optimizer proposed recently \cite{MOE-GAN}.

\subsubsection{Challenge} Although landscape or gradient estimating methods, such as IGL and EGL, take advantage of the universal approximation ability and present encouraging performance on some BBO problems, their inherent natures with local gradient learning determine that they are more capable of solving quasi-convex problems \cite{Sarafian2020EGL,8742787}. Latent manifold learning methods could result in the loss of landscape information owing to inaccurate conversion, though they require weak \emph{a priori} assumption.

The GAN-based solver is a pioneer of GAN-based optimization, but is almost a \emph{local optimizer}. It only pays attention to the exploitation by gradually dividing the search space, but loses the ability of jumping out of local optimum, to say nothing of balancing E-E trade-off. It also has many other weaknesses, e.g., non-smoothly reshaping, premature convergence, and dependency on initial state (see Appendix B for more details). Thus, a broad-spectrum \emph{global optimizer} reshaping distribution flexibly and progressively with strategies of balancing E-E trade-off, is highly desired.

\section{Methodology}
\label{sect-Method}

\input{table_tex/tab_notions.tex}

\subsection{Formulation}
\label{Formulation_label}
Generally, proper definition of $p(\boldsymbol{x})$ guarantees searching efficiency (see Appendix  A). Since there is no meaningful information about the black box $f$, we can only use historical $\boldsymbol{x}$ and $f(\boldsymbol{x})$ recorded  to construct $p(\boldsymbol{x})$. We firstly assume the optimum $\boldsymbol{x}^*$ is drawn from $h(\boldsymbol{x})$, i.e, the distribution of \emph{historical best} solutions $\boldsymbol{x}_\mathrm{opt}$. Refining $h(\boldsymbol{x})$ toward the delta distribution $\delta _{x^ * }(\boldsymbol{x})$ for \emph{exploitation} can be achieved by updating $\boldsymbol{x}_\mathrm{opt}$ with samples from previous $h(\boldsymbol{x})$.

However, if $\boldsymbol{x}^*$ is out of distribution $h(\boldsymbol{x})$, the optimizer would fail in locating it by exploitation solely, which leads to the necessity of additional \emph{exploration} on the global domain $\Omega$. As the uniformly random search is often viewed as an extreme exploration, we use the mixture distribution of $h(\boldsymbol{x})$ and a multivariate uniform distribution $u(\boldsymbol{x})$ as the final $p(\boldsymbol{x})$, which is defined as
  {
  \begin{equation}
  \label{eq-px}
 p(\boldsymbol{x}) = ({1}/({1 + \lambda})) \cdot h(\boldsymbol{x}) + ({\lambda}/({1 + \lambda})) \cdot u(\boldsymbol{x}),
  \end{equation}
  }%
  where the E-E trade-off is tuned by adjustment factor $\lambda$. The $u(\boldsymbol{x})$ is defined as
  {
  \begin{equation}
  \label{eq-udist}
  u\left( \boldsymbol{x} \right) = \left\{ {\begin{array}{*{20}{c}}
  {\frac{1}{{v\left( \Omega \right)}}}&{\boldsymbol{x} \in \Omega }\\
  0&{\boldsymbol{x} \notin \Omega }
  \end{array}} \right.,
  \end{equation}
  }%
  where $v\left( \Omega \right)$ denotes for the volume (area) of $\Omega$. When $\lambda=0$, $p(\boldsymbol{x})$ degenerates to $h(\boldsymbol{x})$. On the contrary, $\lambda=+\infty$ means all positions in $\Omega$ have the same chance to be $\boldsymbol{x}^*$. The $u(\boldsymbol{x})$ can flatten $p(\boldsymbol{x})$ and increase the structural diversity of candidate solutions. Such a design facilitates the recovery of solution diversity lost caused by the refinement of $h(\boldsymbol{x})$ and explores new solutions.

\setcounter{footnote}{1}\footnotetext{Table \ref{tab:notations} lists the notations used in the paper.}

\subsection{Framework of OPT-GAN} 
\label{methodology_A}
\emph{How to achieve $p(\boldsymbol{x})$ in practice?}
Although reshaping $p(\boldsymbol{x})$  has the potential to approach optimum $\boldsymbol{x}^*$, this distribution could be arbitrary shape in high-dimensional space during optimization. Although there are many candidate models for constructing a distribution, the GAN is considered an effective universal distribution learner \cite{Lu2020UAT, Grover_Dhar_Ermon_2018} for multi-dimensional and arbitrary distributions. Thus, in this work, we design a GAN-based framework to carry out the formulation of Eq. \ref{eq-px}.

The OPT-GAN consists of three components: solution generator $G$, bi-discriminators $D=\{D_{I}, D_{R}\}$, and historical best solution set $\boldsymbol{x}_\mathrm{opt}$ (see Fig. \ref{fig:flowchart} ). $G$ learns the mixture distribution $p(\boldsymbol{x})$ by adversarially learning against $D_{I}$   and $D_{R}$ at the same time. $\boldsymbol{x_\mathrm{opt}}$ serves as a knowledge base related to $h(\boldsymbol{x})$ and is continually maintained during optimization. The loss of $G$ exquisitely contains the correction to $h(\boldsymbol{x})$ by $D_{I}$ and correction to $u(\boldsymbol{x})$ by $D_{R}$, gradually enforcing $G$ to approach their mixture distribution $p(\boldsymbol{x})$, provided proper training of $D_{I}$ and $D_{R}$. 

In addition, the mode collapse problem \cite{li2021tackling} may lead to optimization failure. Although refining distribution toward $\boldsymbol{x}^*$ can be viewed as a ``positive" mode collapse, the abnormal collapse during optimization still results in premature loss of solution diversity. Therefore, Wasserstein GAN with gradient penalty (WGAN-GP) \cite{Gulrajani2017WGAN} is adopted to prevent premature convergence to local optima.

  \subsubsection{Exploitation Discriminator}
  \label{D1}
  $D_{I}$ is for exploitation by differentiating generated solutions by $G$ from $\boldsymbol{x}_\mathrm{opt}$, i.e., $h(\boldsymbol{x})$.  It is fed a solution and outputs a value indicating which distribution the solution comes from.\frenchspacing Its loss $\mathcal{L}_\mathrm{D_{I}}$ is defined as
  {
  \begin{equation}
  \label{eq-LossD-Exploitation}
  \begin{split}
  \mathcal{L}_\mathrm{D_{I}}= & \mathbb{E}_{\eta}{[ D_{I}(G(\boldsymbol{\eta}))]} - \mathbb{E}_{\boldsymbol{x}_\mathrm{samp}}{[ D_{I}(\boldsymbol{x}_\mathrm{samp})]} \\
  + & \beta \ \mathbb{E}_{\hat{x_I}} [( \left\| \nabla_\mathrm{\boldsymbol{\hat{x_I}}} D_{I}(\boldsymbol{\hat{x_I}}) \right\|_2 - 1)^2 ],
    \end{split}
  \end{equation}
  }%
where $\boldsymbol{x}_\mathrm{samp}$ is sampled from $\boldsymbol{x}_\mathrm{opt}$ (bootstrapping method). $\boldsymbol{\eta}$ is the random noise, and $\beta$ is gradient penalty factor.
$\boldsymbol{\hat{x_I}}$ is sampled uniformly along straight lines between $G(\boldsymbol{\eta})$ and $\boldsymbol{x}_\mathrm{samp}$ \cite{Gulrajani2017WGAN}.

  \subsubsection{Exploration Discriminator}
  \label{exploration-discriminator}
  $D_{R}$ aims at differentiating sampled solutions by $G$, from samples of $u(\boldsymbol{x})$. It is associated with OPT-GAN's exploration ability. Similar to $D_I$, it outputs a value indicating which distribution a solution draws from. Its loss $\mathcal{L}_\mathrm{D_{R}}$ is defined as 
  {
  \begin{equation}
  \label{eq-LossD-Exploration}
  \begin{split}
  \mathcal{L}_\mathrm{D_{R}}= &
  \mathbb{E}_{\eta}{[D_{R}(G(\boldsymbol{\eta}))]} -\mathbb{E}_{\boldsymbol{x}_\mathrm{uni}}{[ D_{R}(\boldsymbol{x}_\mathrm{uni})]} \\
  + & \beta \  \mathbb{E}_{\hat{x_R}}{[( \left\| \nabla_\mathrm{\boldsymbol{\hat{x_R}}} D_{R}(\boldsymbol{\hat{x_R}}) \right\|_2 - 1)^2]},
  \end{split}
  \end{equation}
  }%
where $\boldsymbol{x}_\mathrm{uni}$ is from $u(\boldsymbol{x})$.  $\boldsymbol{\hat{x_R}}$ is sampled along straight lines between $G(\boldsymbol{\eta})$ and $\boldsymbol{x}_\mathrm{uni}$ \cite{Gulrajani2017WGAN}.

\subsubsection{Solution Generator}
 As a sampler, $G$ generates candidate solutions $\boldsymbol{x}_\mathrm{G}$ from the intrinsic distribution $p(\boldsymbol{x})$ by injecting the noises $\boldsymbol{\eta}$. According to Eq. \ref{eq-px}, $G$ mixes distribution $h(\boldsymbol{x})$ with $u(\boldsymbol{x})$ to balance the E-E trade-off.
  Thus, the generated solutions not only exploit the discovered regions, but also explore the whole searching domain from a global point of view. The loss function $\mathcal{L}_\mathrm{G}$ consists of the adversarial loss against $D_{I}$ and $D_{R}$, which is defined as
  {
  \begin{equation}
  \label{eq-LossG}
  \begin{split}
  \mathcal{L}_\mathrm{G} = \mathbb{E}_{\eta}{[\frac{1}{1+ \lambda} D_{I}(G(\boldsymbol{\eta}))
  + \frac{\lambda}{1+\lambda} D_{R}(G(\boldsymbol{\eta}))]}.
  \end{split}
  \end{equation}
  }

\input{alg.tex}

  \subsection{Knowledge Base Maintaining}
  \label{methodology_C_knowledgebase}
  \emph{How can the optimizer be guided toward the global optimum in the landscape?} 
  Functioning as a knowledge base, $\boldsymbol{x_\mathrm{opt}}$ is a critical component guiding the optimization direction. Solutions in $\boldsymbol{x}_\mathrm{opt}$ are viewed as samples drawn from $h(\boldsymbol{x})$.
  According to Subsection Formulation\ref{Formulation_label}, refining $h(\boldsymbol{x})$ is related to exploitation, and important in reshaping $p(\boldsymbol{x})$ toward the delta distribution $\delta _{x^ * }(\boldsymbol{x})$. In OPT-GAN, $h(\boldsymbol{x})$ is iteratively refined by two strategies: updating and shrinking $\boldsymbol{x}_\mathrm{opt}$.

  \subsubsection{Updating}
  \label{Updating}
Since the information on landscape is stored in historical solutions, updating $\boldsymbol{x}_\mathrm{opt}$ by incoming ones according to their goodness is indispensable to promote $h(\boldsymbol{x})$ at better regions.
The generated solutions in $\boldsymbol{x}_\mathrm{G}$ are firstly evaluated on $f$. Then, $\boldsymbol{x}_\mathrm{opt}$ is replaced by the best $K$ solutions in $\boldsymbol{x}_\mathrm{opt}  \cup \boldsymbol{x}_\mathrm{G}$. This strategy ensures the superiority of $\boldsymbol{x}_\mathrm{opt}$ in the historical solutions, and enables new knowledge flows into $\boldsymbol{x}_\mathrm{opt}$ to refine $h(\boldsymbol{x})$. Actually, the updating strategy enables the model to be attracted by better regions.

  \subsubsection{Shrinking} Ideally, a size-fixing or even size-increasing $\boldsymbol{x}_\mathrm{opt}$ could save more knowledge about the landscape. Especially at the early stage, a large $\boldsymbol{x}_\mathrm{opt}$ improves the diversity of solutions from a global point of view.
  However, as optimization goes on, refining $h(\boldsymbol{x})$ with a large $\boldsymbol{x}_\mathrm{opt}$ could be trapped into a tediously long ``tug-of-war", especially when the problem is multi-modal. Thus, considering the search budget, we design a shrinking strategy to balance E-E trade-off over time.
   This strategy gradually and smoothly reduces the size of $\boldsymbol{x}_\mathrm{opt}$ to push OPT-GAN to focus on the interest region. 
  The size $K$ of $\boldsymbol{x}_\mathrm{opt}$
  at the $t^{th}$ number of function evaluations ($FEs$) is defined 
  as follows 
  {
  \begin{equation}
  \label{eq-adjust-capacity}
  K^{(t)} = \lceil {K^{(0)}} ^{(1- a \cdot (t/MAXFes))} \rceil ,
  \end{equation}
  }%
  where $K^{(0)}$ denotes the initial size of $\boldsymbol{x}_\mathrm{opt}$, $MAXFes$ is the maximum of $FEs$, and $a$ is the shrinking rate. Specifically, $a=0$ means a size-fixing $\boldsymbol{x}_\mathrm{opt}$, and $a\ge1$ means only the best solution stays at the end of optimization (Appendix C includes visualization of the effects of $a$).

  \subsection{Optimization}
  \label{methodology_D_opt}
  \emph{How do we use OPT-GAN to optimize a black-box problem?} The optimization process consists of two successive stages: generator pretraining and distribution reshaping 
  (see Algorithm \ref{alg-OPTGAN} for pseudocode).
  
  \subsubsection{Generator Pre-training}
  Facing a black-box problem, an optimizer should start with an unbiased $p(\boldsymbol{x})$ over the $\Omega$ to have a global field-of-view for avoiding premature convergence, i.e., $p(\boldsymbol{x})$ needs to be $u(\boldsymbol{x})$ initially. However, the intrinsic $p(\boldsymbol{x})$ in a randomly initialized generator hardly meets this requirement.
  Thus, $G$ is pre-trained only with $D_{R}$ for initialization (without consuming $FEs$). 
  During this stage, the loss of $D_R$ is still Eq. \ref{eq-LossD-Exploration}, but the loss of $G$ is redefined as
{
  \begin{equation}
  \label{eq-LossG-pretraing}
  \mathcal{L}_\mathrm{G} = \mathbb{E}_{\eta}{[D_{R}(G(\boldsymbol{\eta}))]}.
  \end{equation}
}

\subsubsection{Distribution Reshaping} After generator pre-training, we start to search for the global optimum in $\Omega$. The solution generator $G$ with intrinsic $p(\boldsymbol{x})$ generates candidate solutions at each epoch (epoch denotes a single period in which OPT-GAN reshapes $p(\boldsymbol{x})$ and generates new candidates). It is for updating $\boldsymbol{x}_\mathrm{opt}$ that will be learnt by $G$ at next step. Ideally, the initial $p(\boldsymbol{x})$ is expected to be reshaped gradually towards the { delta distribution} $\delta _{x^*}(\boldsymbol{x})$. 

In each epoch, $G$ adversarially learns current $p(\boldsymbol{x})$ against $D_{I}$ and $D_{R}$, under guidance of $\boldsymbol{x}_\mathrm{opt}$ and samples from $u(\boldsymbol{x})$. Then, $\boldsymbol{x}_\mathrm{opt}$ is maintained by updating and shrinking.  This process is performed epoch by epoch to reshape $p(\boldsymbol{x})$ progressively. The found best solution is in the last $\boldsymbol{x}_\mathrm{opt}$. Note that training GAN is usually commented as a difficult task. Thus, to stabilize the estimation of current $p(\boldsymbol{x})$, $G$, $D_{I}$, and $D_{R}$ are sufficiently trained at each epoch.

\input{fig_tex/fig_distribution_process.tex}
\input{fig_tex/fig_trandional_compare.tex}

\section{Experiments}
\label{sect-Experiments}
We examine the efficacy of OPT-GAN in terms of its broad-spectrum global optimization ability. The baseline methods are tested on the challenging black-box benchmarks from the COCO platform \cite{hansen2020cocoplat}, CEC'19 Benchmark Suite \cite{price2018problem}, Conformal Bent Cigar \cite{Liu2020Bending}, and Simulationlib \cite{simulationlib}. We also validate the efficacy on two real-world problems.  Considering this study focuses on NN-based broad-spectrum optimizer, two groups of optimizers are adopted.
The first group includes various broad-spectrum optimizers, including BFGS, CMA-ES, {Nelder-Mead method (NM)}, PSO, Bayesian Optimization by Density-Ratio Estimation (BORE) \cite{BORE}, Bayesian Optimization of Risk Measures (BoRisk) \cite{BoRisk}, and BoRisk with 30 candidates (BoRisk-pop30; see Appendix  D.1). The second group includes state-of-the-art NN-based optimizers, Explicit Gradient Learning (EGL), GAN-based Solver (GBS), and Weighted Retraining (WR). {
In order to verify the framework’s effectiveness and eliminate the interference of architectural tricks to evaluation results, we only use compact architectures (see Appendix D.1) to implement the prototypes of the generator and discriminators.}

The detailed configurations about model specifications, benchmarks, and experimental settings are introduced in Appendix D.1, respectively. We also conducted the ablation analysis, convergence analysis, and hyperparameter analysis for OPT-GAN (see Appendix D.2, D.3, and D.4). 

\subsubsection{Visualization of the Learned Distribution}
To evaluate the ability of distribution estimation, Fig. \ref{fig:DistributionProcess_all} visualizes the learnt $p(\boldsymbol{x})$ of OPT-GAN, CMA-ES, and BoRisk (see Appendix  E) during optimization on Gallagher's Gaussian 101-me Peaks function. CMA-ES builds the distribution by Gaussian prior directly, but easily misses the global optima at the later stage of optimization. The internal $p(\boldsymbol{x})$ in BoRisk presents a distribution with spikes because of the maximization of acquisition function. Although it tries to sample few thoughtful solutions at each epoch to save $FEs$, unfortunately, due to the cumulative error effect, BoRisk sampling may miss the optimal regions in the complex landscape. By contrast, OPT-GAN provides redundancy on multiple potential optimal regions via arbitrary shaped distribution, to obtain the holistic view of the problem. Benefiting from a delicate balance between exploration and exploitation on learning $p(\boldsymbol{x})$, OPT-GAN gradually samples some promising solutions and converges to global optima.

\input{fig_tex/fig_nn_dim40_ecdf_hisg}
\input{fig_tex/fig_statistic_ave_rank.tex}

 \subsubsection{Comparison with Traditional Optimizers}
To verify the adaptability to diverse problems, the convergence curves and empirical cumulative distribution function (ECDF) \cite{hansen2020cocoplat} of OPT-GAN are compared with various traditional optimizers (see Fig. \ref{fig:converg-trandional}). Considering an optimizer may have significantly different performance between different benchmark suits, we choose diversified benchmarks from different suits to display the influence of \emph{a priori} assumption between different optimizers.
As the available dimensions between benchmark suites differ, for consistency and fairness, the dimensionality of benchmarks is set to the shared available dimensions, i.e., 2 and 10.
The Sphere has significant Gaussian characteristics suitable to BFGS, NM \cite{schraudolph2007stochastic,gao2012implementing}, and CMA-ES. Unsurprisingly, their performance on the Sphere outperforms OPT-GAN. CMA-ES also shows outstanding performance on the 10-dimensional (10D) Lunacek bi-Rastrigin (local Gaussian-type), though it is worse than OPT-GAN on the 2-dimensional (2D) case. These imply that a successful matching between landscape characteristics and model prior can accelerate optimization.

However, for benchmarks without significant Gaussian shape, OPT-GAN gains ground on competitors. Although PSO performs better on the 2D Shifted and Rotated Weierstrass and 10D Michalewicz, its overall performance is inferior to OPT-GAN. BoRisk and BORE gradually lags behind OPT-GAN as the dimensionality increases. Overall, OPT-GAN shows the best or second-best performance in most benchmarks, reflecting its \emph{broad-spectrum} adaptability.  It can  adapt to diversified landscapes, benefiting from GAN's universal distribution learning ability \cite{huster_pareto_2021}.

In terms of ECDF, although the proportion of trials (PT) of OPT-GAN increases slowly at the early stage, it shows the highest PT at the later stage, indicating it can find the global optimum with low $FEs$. The $t$-test results (see Appendix  D.5) also reveal the superiority of OPT-GAN over competitors in terms of statistical significance.

Fig. \ref{fig:fig-ave_rank_statistic} shows the results of average ranks versus $FEs$, dimensionality, and fitness distance correlation (FDC), separately. In Fig. \ref{fig:fig-ave_rank_statistic} (a), OPT-GAN illustrates a better average rank than its opponents. Although BoRisk is better than OPT-GAN at the early stage, it is surpassed with the increase of $FEs$, because OPT-GAN searches different basins of attraction by balancing E-E trade-off. In Fig. \ref{fig:fig-ave_rank_statistic} (b), OPT-GAN possesses the best average rank for 2D and 10D cases. The performance of BoRisks decreases as problems' dimensionality rises. In short, OPT-GAN presents the best average rank as $FEs$ or dimension increases.

We also visualize the performance vs. global trend of landscape via fitness distance correlation (FDC) \cite{jones1995FDC}, where a higher FDC means the stronger global trend (i.e., Gaussian property). Optimizers are tested on COCO benchmarks with different FDC values (around 1, 0.6, 0.3, 0). Fig. \ref{fig:fig-ave_rank_statistic} (c) shows the average rank of CMA-ES decreases as FDC decreases due to its Gaussian assumptions. This phenomenon is also found on other benchmarks in the COCO platform. CMA-ES performs better on benchmarks with high FDC values (the first 14 benchmarks in the COCO platform). In contrast, OPT-GAN and BoRisks increase, which means they can adapt to diversified problems. Overall,  OPT-GAN is less sensitive to the Gaussian trend of the problem.

\input{fig_tex/fig_realword.tex}
\input{fig_tex/fig_FM.tex}

\subsubsection{Comparison with NN-based Optimizers}
  Considering NN-based optimizers are commonly tested on the well-known COCO platform \cite{Sarafian2020EGL}, the performance results of OPT-GAN and other NN-based optimizers are also summarized in Fig.\ref{fig:converg-nn-dim40-ecdf-hisg}.
  For the convergence curves in Fig.\ref{fig:converg-nn-dim40-ecdf-hisg} (a), OPT-GAN shows the best performance compared with other NN-based optimizers on the majority of benchmarks. For separable problems (F1-F5) and the problems with conditioning (F6-F14), although the convergence speed of OPT-GAN is slower than those of EGL and GBS, it obtains the best fitness in the most cases because OPT-GAN is encouraged to explore different regions at the early stage and to focus on exploiting the basin of attraction at the later stage. In addition, EGL and GBS are easily trapped into local optimum owing to the lack of exploration for the basin of attraction. Thus, for multi-modal problems with many local optima (F15, F17-F22), OPT-GAN still displays the best results. The above phenomena reflect that OPT-GAN has better global optimization ability by balancing E-E trade-off during optimization. WR is difficult to find the best fitness under the limited resources due to its high time consumption. Thus, it performs poorly on most benchmarks. Moreover, the range of average convergence curves of OPT-GAN is smaller than its opponents on most benchmarks, indicating the more stable convergence process of OPT-GAN.

The results from Fig.\ref{fig:converg-nn-dim40-ecdf-hisg} (b) also show the superiority of OPT-GAN over other NN-based optimizers, especially with the increase of dimensions. Although the PT of OPT-GAN lags behind other competitors at the early stages, it boosts rapidly as the optimization proceeds. This is because the strategies to balance E-E trade-off take time to form a global perspective, and OPT-GAN pinpoints
the global optimum gradually along the basin of attraction. 
In Fig.\ref{fig:converg-nn-dim40-ecdf-hisg} (c), the rank of EGL increases at higher dimensionality, whereas those of GBS and WR decrease, as EGL is mainly a local optimizer, avoiding time-consuming deliberative searching in high-dimensional landscapes. Nevertheless, OPT-GAN still exhibits the highest performance. Moreover, OPT-GAN is significantly better than other models on most COCO benchmarks by observing the t-test results (see Appendix  D.6).

\input{fig_tex/fig_AllDistance.tex}

\subsubsection{Performance on Real-world Problems}
We validate the performance of OPT-GAN by two real-world problems. One of them is the optimization of neural network-based symbolic regresser: NEEP \cite{NEEP}, which is a high dimensional optimization problem with very complex landscape.
Another is Frequency Modulated Sounds Parameter Identification (FMSPI) problem \cite{843494}, which plays an essential role in modern music, e.g., emulation of acoustic musical instruments.

Fig. \ref{fig:converg-realworld} shows the average convergence curves of different optimizers on NEEP with five  datasets and their corresponding 2D projected landscapes \cite{li2018visualizing}. The dimension of NEEP optimization problem is 400 for nico5, nico6, and nico11, and 640 for concrete and energy (see Appendix D.7 for more information of NEEP problem). 
It can be seen that each landscape is highly complex and multi-modal, posing a challenge to optimizers on balancing the E-E trade-off. EGL shows premature convergence on most datasets, as it relies on the gradient. Moreover, BoRisk and WR fail to find the optimum within the limited elapsed time because of their high time consumption. Although CMA-ES, PSO, and GBS display a similar performance with OPT-GAN on the Concrete, OPT-GAN exceeds them on the rest datasets. Actually, OPT-GAN exhibits the best overall performance on all datasets. These phenomena illustrate that OPT-GAN can balance E-E trade-off to search for global optimum on complex real-world landscapes.

Fig. \ref{fig:fig_FM} shows the average convergence curves of different optimizers in the FMSPI problem. This problem is a six-dimensional real-world problem with a complex multi-modal structure and strong variable interdependency. OPT-GAN presents the best performance because it can balance the exploration-exploitation trade-off. WR, BoRisk, PSO, and GBS present similar worse performances. CMA-ES performs worst at the early stage. EGL presents the worst performance since it is easy to premature convergence due to a complex multi-modal landscape.

\subsubsection{Convergence Analysis}
The convergence of OPT-GAN can be analyzed from three perspectives. 
The GAN framework is used to make sure the new solutions are sampled from the mixture distribution $p(\boldsymbol{x})$. 
The Knowledge Base Maintaining strategy ensure the solutions in $\boldsymbol{x}_{\mathrm{opt}}$ could cluster to an optimum. 
The uniform distribution $u(\boldsymbol{x})$ promise the global optimum could be sampled out even if it is not covered by the $h(\boldsymbol{x})$.

To verify them, a series of experiments were carried out via several benchmarks. Fig. \ref{fig:fig_AllDistance} (a) shows the Wasserstein estimate curves related to $G$ and $D=\{D_I,D_R\}$. The blue curve gradually fluctuates down and converges to zero gradually; the red curve fluctuates up and converges to a stable value. That means the distribution represented by $G$ can converge to a stable state over time.
Fig. \ref{fig:fig_AllDistance} (b) displays the mean nearest neighbor distance (MNND) between generated solutions. The orange curves gradually converge to zero from the large value as the iteration continues, illustrating that the generated solutions spread in different regions at the early stage and gradually converge together at the later stage.
Fig. \ref{fig:fig_AllDistance} (c) show shows the mean global optimum distances (MGOD) based on $\boldsymbol{x}_\mathrm{G}$ and $\boldsymbol{x}_\mathrm{opt}$, respectively. Both curves start with a large value; then, they fluctuate down and stabilize to a small value. That means the generated solutions could converge toward the region with global optimum. For more information, please refer to Appendix D.3.

\subsubsection{Hyperparameter Analysis}
To analyze the effect of the hyperparameters of OPT-GAN and to explain how to tune them in practice, the main hyperparameters are analyzed and reported in Fig. \ref{fig:converg-para}. Several benchmarks with different preferences for exploration and exploitation ability are used. As a regulator in E-E trade-off, the large $\lambda$  stresses exploring complex landscapes, e.g. multi-modal ones, whereas smaller value promotes exploiting the current attraction basin. The $K$ and $a$ balance the E-E trade off along the timeline. Small $K$ and large $a$ accelerate the transition from exploration to exploitation. See Appendix D.4 for more details.

\subsubsection{Limitations} {We admit that OPT-GAN shares the same disadvantage with other NN-based models: the learning process is time-consuming. It may not be the best to efficiently solve a Gaussian-shape or unimodal problem due to its weak prior knowledge. However, to deal with massive types of real problem features \cite{diversity1, diversity2}, the optimizers with less prior such as OPT-GAN, are highly desirable to a broad extent.
In addition, OPT-GAN also presents comparable time consumption with other NN-based optimizers (see Appendix D.8). }

\input{fig_tex/fig_para.tex}
 
\section{Conclusion}
\label{sect-Conclusion}
This study proposes a \emph{broad-spectrum} global optimizer, named OPT-GAN, for diversified BBO problems. It consists of three collaborative components: solution generator, bi-discriminators, and knowledge base. Adversarial learning between the generator and bi-discriminators enables shaping arbitrary distributions, capturing diversified features in a relatively broad domain. It balances E-E trade-off by three organic and indispensable strategies: supervision of bi-discriminators, updating and shrinking of knowledge base, and pre-training of generator. Experiments show that 
it outperforms various BBO methods, including neural network-based ones, on diversified  problems.

In the future, as we only used the most compact architecture as a prototype of OPT-GAN in the experiment, the network structure requires further attention to take full advantage of GAN's distribution learning ability. In addition, the landscape/problem-related computation methods or adaptive techniques for hyperparameters need to be studied. The extension of the application domain to various real-world problems will also be a concern.

\section*{Acknowledgements}
This work was supported by National Natural Science Foundation of China under Grant No. 61872419, No. 62072213, No. 61873324, No. 61903156. Shandong Provincial Natural Science Foundation No. ZR2022JQ30, No. ZR2022ZD01, No. ZR2020KF006. Taishan Scholars Program of Shandong Province, China, under Grant No. tsqn201812077. “New 20 Rules for University” Program of Jinan City under Grant No. 2021GXRC077.

{
\small 
\bibliography{ref}
}

\clearpage
\includepdfmerge{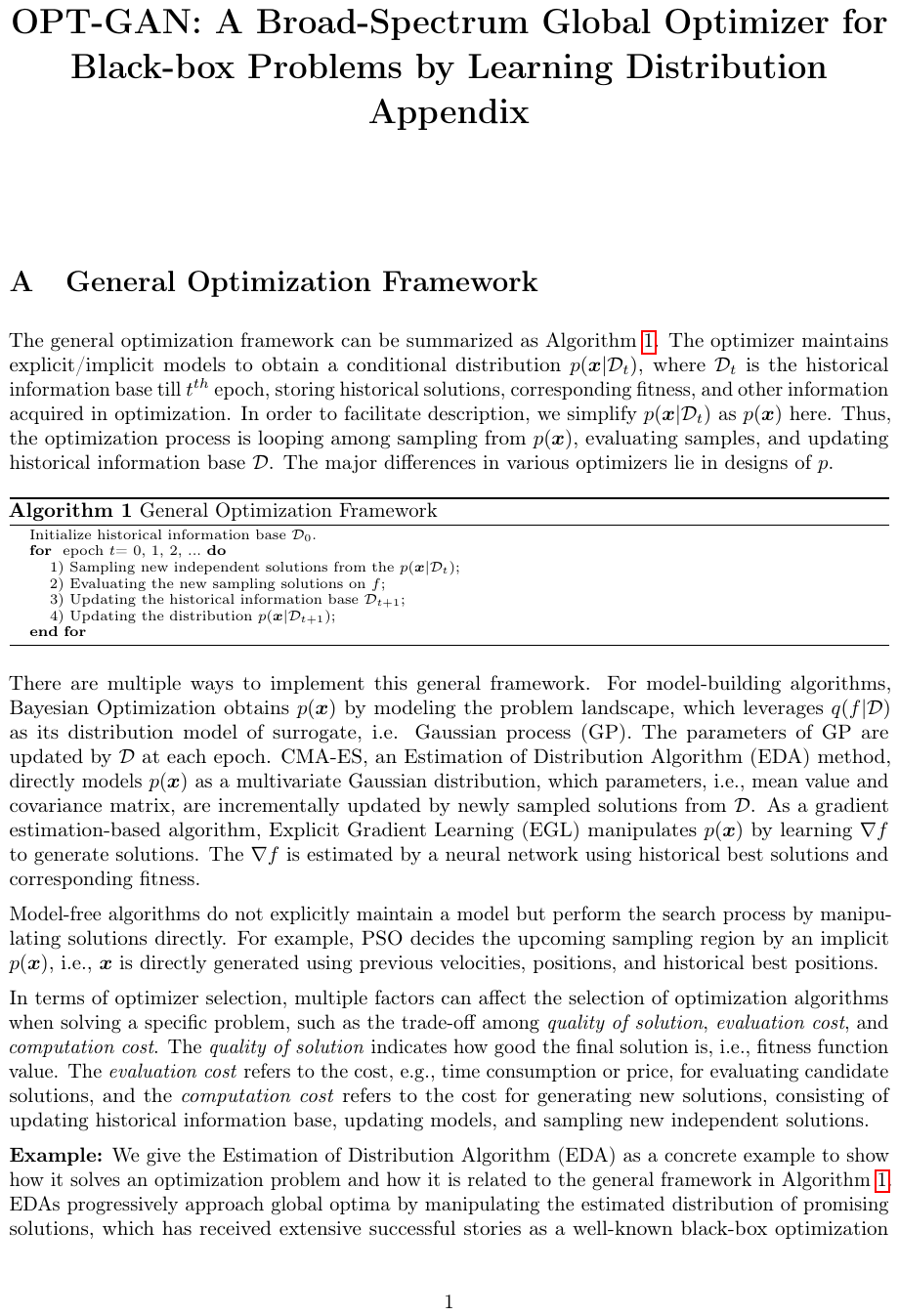,-}

\end{document}


\maketitle

\section{General Optimization Framework}
\label{Preliminary}
The general optimization framework can be summarized as Algorithm \ref{General_optimization_framework}.
The optimizer maintains explicit/implicit models to obtain a conditional distribution $p(\boldsymbol{x}|\mathcal{D}_t)$, where $\mathcal{D}_t$ is the historical information base  till $t^{th}$ epoch, storing historical solutions, corresponding fitness, and other information acquired in optimization. In order to facilitate description, we simplify $p(\boldsymbol{x}|\mathcal{D}_t)$ as $p(\boldsymbol{x})$ here. Thus, the optimization process is looping among sampling from $p(\boldsymbol{x})$, evaluating samples, and updating historical information base $\mathcal{D}$.
The major differences in various optimizers lie in designs of $p$.

\input{supplemental/table_tex/opt_process}

There are multiple ways to implement this general framework. For model-building algorithms, Bayesian Optimization obtains $p(\boldsymbol{x})$ by modeling the problem landscape, which leverages $q(f|\mathcal{D})$ as its distribution model of surrogate, i.e. Gaussian process (GP). The parameters of GP are updated by $\mathcal{D}$ at each epoch. 
CMA-ES, an {Estimation of Distribution Algorithm (EDA)} method, directly models $p(\boldsymbol{x})$ as a multivariate Gaussian distribution,
which parameters, i.e., mean value and covariance matrix, are incrementally updated by newly sampled solutions from $\mathcal{D}$.
As a gradient estimation-based algorithm, {Explicit Gradient Learning (EGL)} manipulates $p(\boldsymbol{x})$ by learning $\nabla f$ to generate solutions. 
The $\nabla f$ is estimated by a neural network using historical best solutions and corresponding fitness.

Model-free algorithms do not explicitly maintain a model but perform the search process by manipulating solutions directly. For example, PSO decides the upcoming sampling region by an implicit $p(\boldsymbol{x})$,
i.e., $\boldsymbol{x}$ is directly generated using 
previous velocities, positions, and historical best positions.

{
In terms of optimizer selection, multiple factors can affect the selection of optimization algorithms when solving a specific problem, such as the trade-off among \emph{quality of solution}, \emph{evaluation cost}, and \emph{computation cost}. The \emph{quality of solution} indicates how good the final solution is, i.e., fitness function value. The \emph{evaluation cost} refers to the cost, e.g., time consumption or price, for evaluating candidate solutions, and the \emph{computation cost} refers to the cost for generating new solutions, consisting of updating historical information base, updating models, and sampling new independent solutions.
}

\textbf{Example:} {We give the Estimation of Distribution Algorithm (EDA) as a concrete example to show how it solves an optimization problem and how it is related to the general framework in Algorithm \ref{General_optimization_framework}. EDAs progressively approach global optima by manipulating the estimated distribution of promising solutions, which has received extensive successful stories as a well-known black-box optimization algorithm family\cite{NIPS1996_MIMIC, EDA_TEVC2, HAUSCHILD2011111, Hansen2003CMAES}. In EDA, $\mathcal{D}$ corresponds to the good population set, whose functionality is similar with $ \boldsymbol{x}_\mathrm{opt} $ in Algorithm \ref{alg-OPTGAN}. 
It constructs $p(\boldsymbol{x})$ based on $\mathcal{D}$ with a parametric distribution model, e.g., the Gaussian model. Note that its $p(\boldsymbol{x})$ only incorporates $h(\boldsymbol{x})$ without considering the explorative role of $u(\boldsymbol{x})$ (see Eq. 2 in paper), which easily leads to over exploitation, especially for continuous EDA methods in solving black-box problems. Although EDAs have been widely applied in many real-world problems, thorough theoretical analysis are still difficult to be conducted since these methods are mainly used to deal with black-box optimization problems \cite{sala_global_2017, 1705405, EDA_theoreticalAnalysis}.
}

\section{Weakness of GAN-based Solver}
The GAN-based solver could search original space directly for arbitrary black-box problems, and is a pioneer of GAN-based optimization. However, it actually has many weaknesses. 
\begin{itemize}[leftmargin = 10 pt, topsep = 0 pt, partopsep = 0 pt]
\item It is almost a \emph{local optimizer}, as it only pays attention to the exploitation by gradually dividing the searching space, but loses the ability of jumping out of local optimum, to say nothing of balancing {exploration-exploitation (E-E)} trade-off. 
\item The use of fixed size and binary knowledge base cannot guide the generator to smoothly tune the distributions on the search domain. 
\item Learning the distribution accurately is difficult because of the mode collapse, which also results in premature convergence. 
\item Considering its nature of exploitatory searching and arbitrary initialization, the success of optimization is largely dependent on the intrinsic distribution of the initial generator. 
\item Only a couple of problems are used for verifying the efficacy, lacking of extensive experiments to demonstrate the features of GAN-based optimization. 
\end{itemize}

\section{Methodology}
{Table \ref{tab:notations} lists the notations used in the Methodology. Algorithm \ref{alg-OPTGAN} is the full version of OPT-GAN's pseudocode. It describes how OPT-GAN is used to optimize a black-box problem. Fig \ref{fig:k_curve} visualizes the influence of $a$ on $K$.}

\input{supplemental/alg.tex}

\section{Experiments}

\subsection{Experiment Setting}

\input{supplemental/table_tex/tab_notations.tex}
\textbf{Benchmark Settings}
In the experiments, challenging black-box benchmarks are employed from the COCO platform \citep{hansen2020cocoplat}, CEC'19 Benchmark Suite \citep{price2018problem}, Conformal Bent Cigar \citep{Liu2020Bending}, and Simulationlib \citep{simulationlib}. The information of these benchmarks is given in Table \ref{benchmarkinfo}.

\textbf{Architecture of OPT-GAN}
{In experiment, the architectures of $D_I$, $D_R$, and $G$ of OPT-GAN are multi-layer perceptrons (MLP) with single hidden layer. The input noise of $G$ is drawn from a multivariate uniform distribution $U[\boldsymbol{-1}, \boldsymbol{1}]$ with dimensionality $2n$, where $n$ is the benchmark's dimensionality. The number of neurons in the hidden layer is 50 for the generator and discriminators. LeakyRelu is served as the activation function.}

\input{supplemental/fig_tex/fig_K.tex}

\input{supplemental/table_tex/benchmarks_inf.tex}

{Please note that the architecture of the generator and discriminators in the experiment is by no means the best implementation of OPT-GAN. However, it is one of the most straightforward implementations to verify the framework's effectiveness, eliminating the interference of architectural tricks to evaluation results.}

\textbf{Hyperparameter Settings}
The main hyperparameter settings of these optimizers are described as follow.
For a fair comparison, the hyperparameters of OPT-GAN and {GAN-based Solver (GBS)} are obtained by trial and error.
We take the recommendations \citep{Sarafian2020EGL} for BFGS, {Nelder-Mead method (NM)}, CMA-ES, and {Explicit Gradient Learning (EGL)}. For PSO, the settings of hyperparameters follow the work \citep{tang2021novel}. For {Weighted Retraining (WR)}, the hyperparameters are set according to the Ref.\citep{NEURIPS2020_81e3225c}.
{For a fair comparison, we also provide BoRisk-pop30, a variation of BoRisk with identical configurations to it \citep{BoRisk}, except for the candidate set size and the initial set size.}
{The detailed hyperparameter settings of OPT-GAN and opponents are described in Table \ref{tab:ParameterSettings}, respectively.} Python is used to code these optimizers. {Meanwhile, in order to maintain the fairness of all the experiments, we ran all algorithms on the CPUs, even though the NN-based methods can be accelerated on the high-performance platform.}
   
The indicator, $fbest-f(\boldsymbol{x}^*)-prec$ \citep{hansen2020cocoplat}, is used to record the fitness of optimizers, where $fbest$ is the historical best fitness, and $prec=10^{-8}$ denotes the predefined precision level. The lower fitness on the benchmarks indicates the better performance of optimizers.Unless otherwise noted, each optimizer runs 15 repeated trials on each problem. When $n=2$, the $MaxFes$ is 50000. On the other dimensions, the $MaxFes$ is 100000. 
A single run stops if: (1) fitness reaches the precision ($fbest-f(\boldsymbol{x}^*)-prec<0$), (2) $FEs$ arrives at the maximum number, or (3) the elapsed time exceeds 3 hours. 

{The proportion of trials (PT) of empirical cumulative distribution function(ECDF\footnote{The empirical (cumulative) distribution function $F : \mathbb{R} \rightarrow [0,1] $ is defined for a given set of real-valued data $S$, such that $F(x)$ equals the fraction of elements in $S$ which are smaller than $x$. The function $F$ is monotonous and a lossless representation of the (unordered) set $S$.}) for an optimizer equals the number of trials whose fitness reaches the precision ($fbest - f^* > 10^{1}$, i.e. successful) divided by the total number of trials.}

\input{supplemental/table_tex/tab_exp_setting.tex}
\subsection{Ablation Analysis}
\input{supplemental/fig_tex/fig_ablation}
 To examine the necessity of different components of OPT-GAN, we conduct the ablation experiments for exploitation discriminator $D_I$, exploration discriminator $D_R$, shrinking strategy, generator pre-training, and bi-discriminators component. Fig.  \ref{fig:ablation} visualizes the distribution learned by the generator $G$ during optimization in the way of the heatmap.
  No-$D_I$ OPT-GAN is an OPT-GAN without the exploitation discriminator $D_I$; No-$D_R$ OPT-GAN is an OPT-GAN without the exploration discriminator $D_R$; No-Shrinking OPT-GAN represents an OPT-GAN without the shrinking strategy; No-Pre-train OPT-GAN denotes an OPT-GAN without the generator pre-training.
  { {Single-D OPT-GAN is an OPT-GAN that replaces the bi-discriminators with the single discriminator for a mixture of $h(\boldsymbol{x})$ and $u(\boldsymbol{x})$ to learn $p(\boldsymbol{x})$.}}
 
  \textbf{Ablation Analysis for $D_{I}$ and $D_{R}$}
  The experiments of No-$D_{I}$ OPT-GAN and No-$D_{R}$ OPT-GAN are used to verify the influence of $D_{I}$ and $D_{R}$, respectively. $D_{I}$ determines the exploitation, and $D_{R}$ contributes to the exploration. When $D_I$ is turned off, the solution generator $G$ only learns from the $u(\boldsymbol{x})$ . No-$D_I$ OPT-GAN is similar to a random search optimizer given the lack of the ability to focus on the basin of attraction. Thus, the distribution learned by No-$D_I$ OPT-GAN cannot converge to the optimum in Fig. \ref{fig:ablation} (a). On the contrary, when $D_R$ is turned off, the solution generator $G$ only learns from the knowledge base, and the mixture distribution $p(\boldsymbol{x})$ denoted by $G$ is equal to $h(\boldsymbol{x})$. In Fig.  \ref{fig:ablation} (b), although the pre-training strategy initializes $G$ with a global viewpoint, No-$D_R$ OPT-GAN rapidly converges to a local optimum because it loses the exploration ability and degenerates into a local optimizer.

\textbf{Ablation Analysis for Shrinking Strategy}
No-shrinking OPT-GAN is designed to show the influence of the shrinking strategy {which controls the shrinking rate of $K$ (size of $\boldsymbol{x}_\mathrm{opt}$).} According to the distributions learned by No-Shrinking OPT-GAN and OPT-GAN in Fig. \ref{fig:ablation} (c) and (f), respectively, both can learn a nice ``arc" which represents the best region. However, OPT-GAN exploits this ``arc" region faster than No-shrinking N at the later stage and converges to the optimum gradually,{ which due to OPT-GAN can quickly refine $h(\boldsymbol{x})$ (for exploitation) when $K$ is small.} The above phenomena reflect that the shrinking strategy reinforces OPT-GAN's exploitation ability over time and benefits its convergence.

\textbf{Ablation Analysis for Generator Pre-training}
  Generator pre-training initializes the {$G$} and endows $G$ with a global field-of-view for generating uniformly scattered solutions initially in the searching domain. No-Pre-train OPT-GAN is adopted to demonstrate the significance of generator pre-training. In Fig. \ref{fig:ablation} (d), the search regions covered by No-Pre-train OPT-GAN only occupy a part of the search domain at initialization. Although No-Pre-train OPT-GAN explores more regions over time {owing to the support from $D_{R}$}, it rapidly converges to the local optimum compared with OPT-GAN. These phenomena illustrate that OPT-GAN initialized with a global field-of-view is beneficial to avoid premature convergence and find optimum.

\textbf{Ablation Analysis for Single-D OPT-GAN}
  {Single-D OPT-GAN is designed to illustrate the significance of the bi-discriminators. Fig.  \ref{fig:ablation} (e) shows the distribution learned by Single-D OPT-GAN. Both OPT-GAN and Single-D OPT-GAN generators estimate the mixture distribution of $h(\boldsymbol{x})$ and $u(\boldsymbol{x})$. As one component of this mixture distribution is the uniform distribution, the generator $G$ is more interested in the region where $h(\boldsymbol{x})$ gives a higher probability to sample than other regions. For Single-D OPT-GAN, the single discriminator does not explicitly guide $G$ to balance the exploitation ($h(\boldsymbol{x})$) and exploration ($u(\boldsymbol{x})$) in the learning process simultaneously, $G$ may prefer to estimate $h(\boldsymbol{x})$ more accurately in order to decrease adversarial loss. Thus, Single-D OPT-GAN may focus on exploitation rather than exploration over time. It can be observed that Single-D OPT-GAN rapidly converges to a local optimum. For OPT-GAN, $G$ is guided by $D_I$ and $D_R$,  respectively, and is enforced to confuse $D_I$ and $D_R$ simultaneously. Thus, $G$ needs to consider the balance of exploration ($u(\boldsymbol{x})$) and exploitation ($h(\boldsymbol{x})$) to decrease its adversarial loss, because the generator in the GAN with multi-discriminators tends to prefer the balance for all discriminators \citep{pmlr-v97-albuquerque19a}. Although the convergence speed of OPT-GAN is slower than Single-D OPT-GAN, OPT-GAN converges to the global optimum gradually. These phenomena reveal the performance of OPT-GAN with bi-discriminators is better than the single discriminator. }

\input{supplemental/fig_tex/fig_AllDistance.tex}
\subsection{Convergence Analysis}
\label{exp_analysis}
{
As discussed in Section \ref{Preliminary}, EDA algorithms are one of the commonly used global optimizers, proven by numerous practical applications and experiments \cite{HAUSCHILD2011111, 6557593, NIPS2016_289dff07}.
Unfortunately, the theoretical performance analysis of EDA, as metaheuristic algorithms, is still limited and only restricted to particular problem types. Usually, the experimental study is used to analyze the properties of EDA algorithms, such as convergence analysis\cite{sala_global_2017, 1705405,  EDA_theoreticalAnalysis}.
}

In order to verify the convergence of OPT-GAN, a series of experiments are carried out via several benchmarks from three perspectives, including (a) whether the distribution represented by $G$ can converge to a stable state over time, (b) whether the generated solutions by $G$ can converge together as iteration goes on, and (c) whether the generated solutions converge toward the region with global optimum.

Fig. \ref{fig:fig_AllDistance} (a) provides the Wasserstein estimate curves related with $G$ and $D=\{D_I,D_R\}$. The Wasserstein estimate is proposed in Wasserstein GAN \citep{arjovsky2017wasserstein}, and is used to estimate the Earth-Mover Distance between the distribution denoted by generator and the distribution behind the ``real" samples. When Wasserstein estimates move toward a fixed value, it means that the distribution represented by the generator converges to a stable state.
  
  The Wasserstein estimate between $G$ and $D_I$ ($\mathcal{W}_\mathrm{D_{I}}$) is defined as
  \begin{equation}
  \label{eq-WD-Exploitation}
  \begin{split}
  \mathcal{W}_\mathrm{D_{I}}=
  \frac{1}{S}\sum\limits_{s=1}^S
  ({D_{I}(G(\boldsymbol{\eta}^{(s)})) - D_{I}(\boldsymbol{x}_\mathrm{sampling}^{(s)})}),
  \end{split}
  \end{equation}
  and the Wasserstein estimate between $G$ and $D_R$ ($\mathcal{W}_\mathrm{D_{R}}$) is defined as 
  \begin{equation}
  \label{eq-WD-Exploration}
  \begin{split}
  \mathcal{W}_\mathrm{D_{R}}=
  \frac{1}{S}\sum\limits_{s=1}^S
  ({D_{R}(G(\boldsymbol{\eta}^{(s)})) - D_{R}(\boldsymbol{x}_\mathrm{uniform}^{(s)})}).
  \end{split}
  \end{equation}

  In Fig. \ref{fig:fig_AllDistance} (a), the blue curve that represents $\mathcal{W}_\mathrm{D_{I}}$ fluctuates down and converges to zero gradually; the red curve that represents $\mathcal{W}_\mathrm{D_{R}}$ fluctuates up and converges to a stable value.
  The pre-training strategy aids $G$ to learn a multivariate uniform distribution $u(\boldsymbol{x})$ before optimization. Thus, the initial value of $\mathcal{W}_\mathrm{D_{R}}$ is small, whereas $\mathcal{W}_\mathrm{D_{I}}$ is large. As iteration goes on, $\boldsymbol{x}_\mathrm{opt}$ is constantly updated by the generated solutions from $G$, and its size gradually shrinks until the $K=1$. As a result, the distribution $h(\boldsymbol{x})$ denoted by the optimal set becomes stable gradually, and $\mathcal{W}_\mathrm{D_{I}}$ converges to zero.  $p(\boldsymbol{x})$ is also able to move toward a stable distribution, because $p(\boldsymbol{x})$ is the mixture distribution of $h(\boldsymbol{x})$ and a fixed multivariate uniform distribution $u(\boldsymbol{x})$.
  
{
 Fig. \ref{fig:fig_AllDistance} (b) displays the mean nearest neighbor distance (MNND) between generated solutions at each iteration. The MNND denotes the mean of the Euclidean distances between each generated solution and its nearest neighbor in $\boldsymbol{x}_\mathrm{G}$. It can be observed that the orange curves gradually converge to zero from the large value as the iteration goes on, which illustrates that the generated solutions spread in different regions at the early stage and gradually converge together at the later stage.
  Fig. \ref{fig:fig_AllDistance} (c) shows the mean global optimum distances  based on $\boldsymbol{x}_\mathrm{G}$ and $\boldsymbol{x_{opt}}$, respectively. The mean global optimum distance (MGOD) is the mean of the Euclidean distances between the global optimum  $\boldsymbol{x}^*$ and each solution. The blue curve (or green curve) represents MGOD between $\boldsymbol{x}^*$ and the generated solutions in $\boldsymbol{x}_\mathrm{G}$ (or the historical best solutions in $\boldsymbol{x_{opt}}$). The blue and green curves start with a large value; then, they fluctuate down and stabilize to a small value. This phenomenon reveals that OPT-GAN is able to converge to the global optimum. Therefore, the distribution $p(\boldsymbol{x})$ represented by $G$ can converge to a stable state over time, and move toward the delta distribution whose spike is located at $\boldsymbol{x}^*$, according to Fig. \ref{fig:fig_AllDistance} (a), (b), and (c).
  }
  
\input{supplemental/fig_tex/fig_para.tex}

\subsection{Hyperparameter Analysis}
  \emph{How can the hyperparameters of OPT-GAN be tuned in practice?}
  Although OPT-GAN involves multiple hyperparameters, some (e.g., parameters related to the GAN) are analyzed \citep{miyato2018spectral, kingma2014adam}. Thus, only the main hyperparameters related to the proposed OPT-GAN, which are the adjustment factor $\lambda$, shrinking rate $a$, and optimal set size $K$, are analyzed and reported in Fig.  \ref{fig:converg-para}. 
  Several benchmarks with different preferences to the exploration and exploitation ability are used. The dimensionality is set to 10. The Rastrigin function includes many local optimal regions around the optimum, and is preferred by the optimizer with strong exploitation ability.
  The Rotated Rosenbrock function is an ill-conditioned function with a bending valley, which is preferred by the optimizer with strong exploration ability. Lunacek bi-Rastrigin function possesses two funnel-shaped regions, and the global optimum is located in the smaller funnel, which poses the challenge for the optimizer's ability  to balance E-E trade-off.

  The adjustment factor $\lambda$ guides the trade-off between exploration and exploitation at each iteration by controlling the effects of $h(\boldsymbol{x})$ (for exploitation) and $u(\boldsymbol{x})$ (for exploration) on $p(\boldsymbol{x})$ (for estimation the distribution of optimum). {When $\lambda$ is small, $p(\boldsymbol{x})$ is dominated by $h(\boldsymbol{x})$}, which enables OPT-GAN to display a better exploitation ability. With the increase of $\lambda$, the influence of $u(\boldsymbol{x})$ on $p(\boldsymbol{x})$ increases gradually, enhancing the exploration ability of OPT-GAN. Extremely, if $\lambda$ is $+\infty$, then it degrades into a random search optimizer.
  In Fig. \ref{fig:converg-para}, Rastringin function prefers low $\lambda$ because it stresses more on exploitation ability than on exploration ability.
  For the Rotated Rosenbrock function, OPT-GAN achieves better performance when $\lambda$ is large as this benchmark has two peaks. This characteristic forces an optimizer to explore the searching domain. The same phenomenon can be observed on Lunacek bi-Rsatrigin because of the existence of two major basins of attraction. Thus, a carefully tuned $\lambda$ can balance the exploration and exploitation of OPT-GAN. Balancing E-E trade-off would boost the performance on diversified landscapes. Overall, 
  $\lambda$ is suggested with the small value for the  unimodal problems, and is recommended with the large value for multi-modal optimization problems.
  
  The initial size of the optimal set $K$ and its shrinking rate $a$ are interrelated. $K$ defines the initial size of the optimal set, and   $a$ controls how fast the size of the optimal set reduces. The resultant force of $K$ and $a$ can control the E-E trade-off along the time line. A small $a$ indicates a slow shrinking speed. A large $K$ with a small $a$ supports the optimal set with a large size, and its size decreases slowly, which enables the exploration ability of OPT-GAN to reduce slowly. In Fig.  \ref{fig:converg-para}, with the increase of $K$, the performance improves on the Rotated Rosenbrock, and declines on the Rastrigin. These phenomena reveal that the exploration ability raises as $K$ increases. 
  However, when $a$ is over small or $K$ is over large, refining $h(\boldsymbol{x})$ could be slow and trapped into a ``tug-of-war” as the iteration proceeds, which declines the convergence speed of OPT-GAN. In addition, OPT-GAN performs better on the Rastrigin with the growth of $a$. A large $a$ leads to a rapid decline in the size of the optimal set and boosts the preference to exploitation. {{ Therefore, $K$ and $a$ should be tuned carefully to balance the exploration and exploitation. Overall, $K$ combined with a small $a$ is suggested for the multi-modal problems in order to boost the exploration ability of OPT-GAN, and $K$ combined with a large $a$ is 
 recommended for the problems that are preferred by the optimizer with strong exploitation ability.}}
  
\input{supplemental/table_tex/tab_ttest_trandional.tex}

\input{supplemental/fig_tex/fig_nn_dim2.tex}
\input{supplemental/fig_tex/fig_nn_dim10.tex}

\subsection{Comparison with Traditional Optimizers}
To validate the performance of OPT-GAN compared with traditional optimizers in terms of statistical significance, a t-test is conducted. The results are reported in Table \ref{tab:compare-ttests-trandional}.
``Better", ``Same", and ``Worse" represent the number of benchmarks that the performance of OPT-GAN is significantly better, almost the same as, and significantly worse than the opponents, respectively. 
``Merit" is ``Better" minus ``Worse". It can be observed that OPT-GAN is better than BoRisk, BoRisk-pop30, BFGS, CMA-ES, and Nelder-Mead in 2 and 10-dimensional problems.
Moreover, ``Merit" rises with the increase of the dimensionality compared with PSO, BoRisk and BoRisk-pop30. However, compared with BFGS, Nelder-Mead, and CMA-ES, ``Merit"  decreases as the dimensionality increases from 2 to 10. This is perhaps because the complexity of certain high-dimensional problems results in OPT-GAN having to deliberate on the landscape, slowing down the optimization.

\subsection{Comparison with NN-based Optimizers}
We conduct a t-test to analyze the statistical significance of OPT-GAN over other neural network competitors. The results are reported in Table \ref{tab:compare-ttests}. OPT-GAN is significantly better than other models on most COCO benchmarks. The ``Merit" of OPT-GAN over GBS verifies that the adopted strategies for balancing E-E trade-off improves GAN for global optimization. 
  EGL displays better performance as the dimensionality increases because it is designed to tackle the high dimensional problems \citep{Sarafian2020EGL}; even so, the ``Better" and ``Merit" of OPT-GAN exceed EGL. Furthermore, the ``Merit" of OPT-GAN boosts as the dimensionality increases compared with WR. A possible explanation is that the high time consumption of WR affects its ability to find the optimum under the limited resources.

We also provide all the convergence results of those NN-based optimizers on all the 24 problems in 2D, 10D problems (Fig. \ref{fig:converg-nn-dim2} and Fig. \ref{fig:converg-nn-dim10}). On F3, F15-F18, F21, and F22, OPT-GAN has an undeniable advantage over other optimizers in 2 and 10 dimensions, while in F1, F14, and F23, it is equal to or even slightly worse than EGL. By observing the contour maps of different functions, we deduced that EGL, as a local gradient algorithm, is good at solving unimodal ill-conditioned problems. While OPT-GAN, as a global optimizer, has better average performance and is good at solving multi-modal or non-convex problems.

\input{supplemental/fig_tex/fig_nn_ecdf+hisg}
Fig. \ref{fig:fig_NN_hisg}  displays the comparative results of the ECDF in 2 and 10 dimensions. The results demonstrate the superiority of OPT-GAN over other neural network-based optimizers, especially with the increase in dimensionality. Although the PT of OPT-GAN lags behind other competitors at the early stages, it boosts rapidly as the optimization proceeds. This is because the strategies to balance E-E trade-off take time initially to form a global perspective, and OPT-GAN pinpoints the global optimum gradually along the basin of attraction.

\subsection{Experimental Configuration for Optimizating NEEP Problem}
{In this part, we validate the performance of OPT-GAN by a real-world problem of optimizing the neural network-based symbolic regression method (NEEP)\cite{NEEP}.
As a symbolic regression model, a recurrent neural network (RNN) in NEEP is used to generate the expression trees that describe the relationship between the input variables and the output variable. Therefore, the optimization of NEEP is to optimize the connection weights of RNN, which is a multi-modal optimization problem with the high dimensionality. The mean square error (MSE) can be considered as the fitness function.
In the optimization, the uncertainty of the expression tree, such as structure and length, causes the uncertain evolved function. It is difficult to directly compute gradients with BP for training RNN in NEEP \cite{NEEP}. As an E-E trade-off global optimizer, OPT-GAN can optimize NEEP without considering the gradient.}

We evaluated the performance of OPT-GAN compared with neural network-based optimizers (GBS, WR, EGL), and the traditional optimizers (BoRisk, CMA-ES, PSO) by optimizing NEEP on 3 synthetic symbolic regression problems (nico5, nico6, nico11) \cite{nico} and 2 UCI data sets (Concrete and Energy)\cite{uci}. 
\textbf{The dimension of NEEP optimization problem is 400 for nico5, nico6, and nico11, and 640 for concrete and energy.}
For visualizing the 2-dimensional (2D) projected landscapes \cite{li2018visualizing} of NEEP, we use two random vectors named d1 and d2, with range of [-2, 2]. Precisely, They can map the area of the high-dimensional landscape around the weight optimum into the 2D bounded space of [-2, 2].

The generated expression tree is based on the function symbol set $\{+,-,\times,/,sin,cos,\ln \left| x \right| \}$. In order to tackle that divisor is zero, the $/$ is realized by $x_1/(x_2+\varphi)$ where $\varphi=0.01$;  and the $\ln$ is protected by $\ln \left| x +\varphi \right|$.  The hyperparameters of the optimizers follow the table \ref{tab:ParameterSettings}.
Each optimizer repeats 15 times on each dataset, the maximum number of iterations is 100,000. 
A single run is terminated if: (1) fitness reaches the precision ($fbest-f^*-prec<0$), (2) $FEs$ arrives at the maximum number, or (3) the elapsed time exceeds 10 hours.

\input{supplemental/table_tex/tab_ttest_NN.tex}

\input{supplemental/table_tex/tab_time.tex}

\subsection{Time Consumption of Learning-based Algorithms}
In Table \ref{tab:time-compare-chart}, we present the time consumption of learning-based algorithms. 
OPT-GAN has comparable time consumption with other NN-based optimizers.
BoRisk performs a posteriori evaluation based on all historical candidates on the objective function and models a Gaussian process, which makes its time consumption increase exponentially with $FEs$. Therefore, according to the experimental results, the time consumption of BoRisk at $FEs=10000$ is much larger than 24 hours. 
In contrast, GBS uses the real objective function for evaluation during the neural network training process, so $FEs$ are consumed quickly and with low CPU time consumption. However, this optimization approach makes GBS underfitting with limited $FEs$ and poor accuracy of modeling in optimization (see Fig. \ref{fig:converg-nn-dim2}- Fig. \ref{fig:fig_NN_hisg}).

{
Despite the fact that OPT-GAN, as well as most NN-based optimizers, is more time-consuming than classical optimizers, 
with the development of hardware, such as the increase in computing power, decrease in hardware price, and especially development of application-specific processors (see AI Accelerators \citep{9499913,10.1145/2654822.2541967}), the time-consumption gap between deep learning-based optimizers and classical ones will become negligible. This type of method will then be a promising direction by providing better solutions within a similar total optimization time.}

\subsection{Computing Resource}
The numerical experiments were collected from several machines, including Intel(R) Core(TM) i7-10700K 3.80GHz CPU with Linux operating system, Intel(R) Xeon(R) CPU E5-2620v3 2.40GHz CPU X2 with Linux operating system, Intel(R) Xeon(R) CPU E5-2620 2.00GHz CPU X2 with Linux operating system, AMD Rome 7H12 2.60GHz CPU with Linux operating system, Intel(R) Xeon(R) Silver 4214R 2.40GHz with Linux operating system, Intel(R) Core(TM) i7-8750H 2.20 GHz CPU NVIDIA GeForce GTX 1060Ti GPU with Windows 10 operating system.

\section[Visualization]{Visualization of $\boldsymbol{p(x)}$}
{
The visualization of the distribution $p(\boldsymbol{x})$ contains a true landscape of the problem and a heatmap of $p(\boldsymbol{x})$. The landscape of the problem is visualized as a contour map. 
The heatmap of $p(\boldsymbol{x})$ is approximated by Monte Carlo sampling with one million samples.
Sampling methods from $p(\boldsymbol{x})$ are different in different optimizers due to the various designs of $p$.}

{
According to Section \ref{Preliminary}, CMA-ES explicitly obtains the $p(\boldsymbol{x})$ from a Gaussian model. Thus, the one million solutions are sampled from the updated Gaussian model in a stage.
OPT-GAN obtains the $p(\boldsymbol{x})$ by the intrinsic distribution of $G$, which is used to sample the one million solutions with different input random noise.
BoRisk, as a surrogate optimization algorithm, indirectly learns $p(\boldsymbol{x})$ by GP, which models the problem landscape $q(f)$. BoRisk generally samples a candidate solution from $p(\boldsymbol{x})$ by maximizing the acquisition function over the GP. The one million solutions are obtained by randomly repeating the process of optimizing the acquisition function one million times.
}

\bibliographystyle{unsrtnat}
\bibliography{ref_supp}

%% file: fig_tex/fig_sketch.tex
\begin{figure}[t]
    \centering
    \setlength{\belowcaptionskip}{-11 pt}
    \includegraphics[width=0.99 \columnwidth]{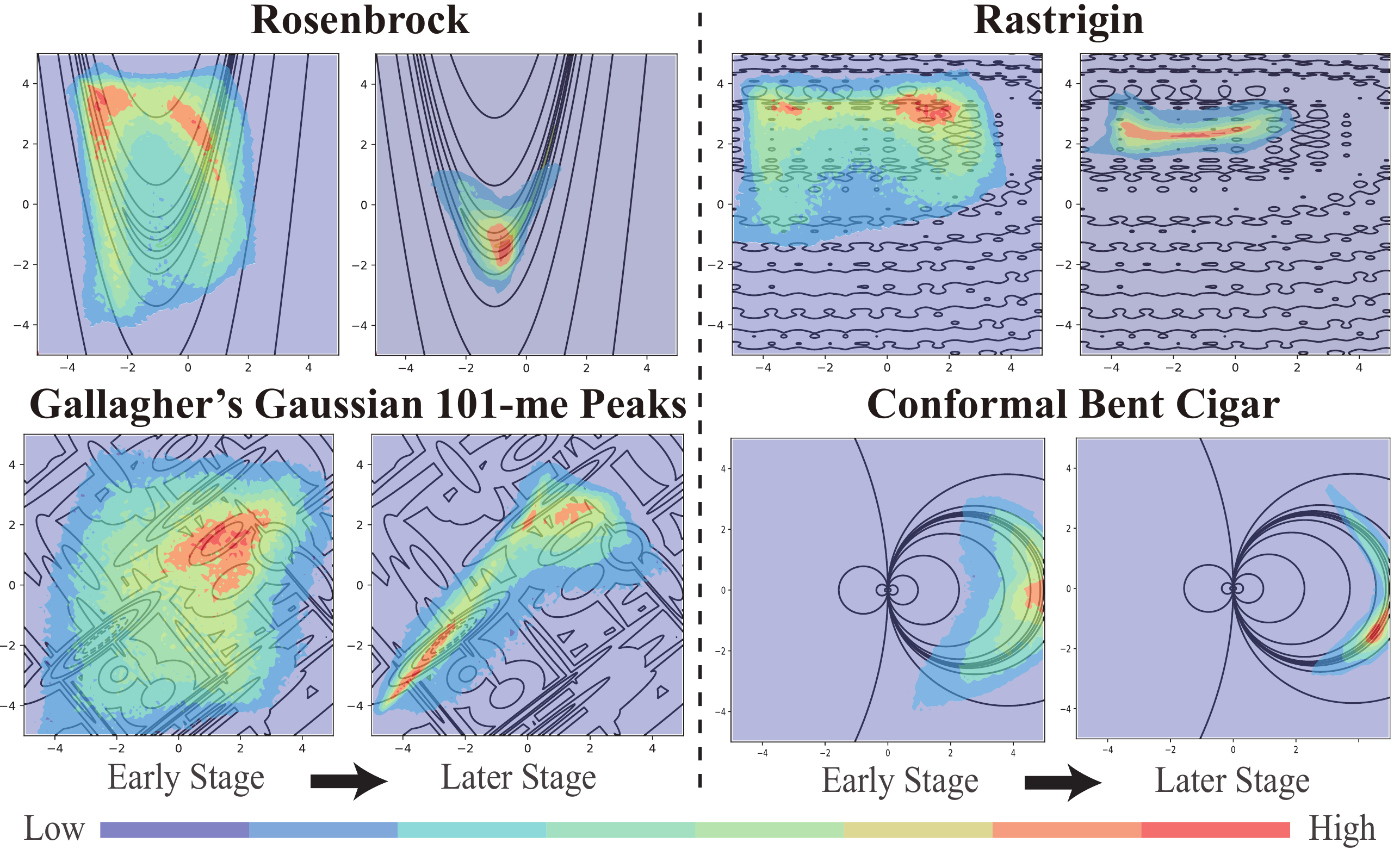}
    \caption{Estimated $p(\boldsymbol{x})$ using OPT-GAN on four problems (see Appendix E for the plotting details of $p(\boldsymbol{x})$). It reshapes $p(\boldsymbol{x})$ progressively according to properties of a landscape.}
    \label{fig:schematic}
\end{figure}

%% file: fig_tex/fig_flowchart.tex
\begin{figure*}[t]
  \centering
    \setlength{\belowcaptionskip}{-9 pt}
  \includegraphics[width=0.9 \textwidth]{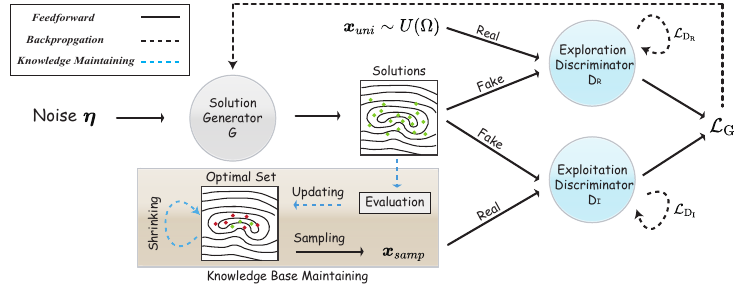}
  \caption{
  The framework of the proposed OPT-GAN. The $\emph{U}(\Omega)$ is a uniform
distribution on $\Omega$.
  } 
  \label{fig:flowchart}
\end{figure*}

%% file: table_tex/tab_notions.tex
\begin{table*}[ht]
\centering
\caption{Description of notations.}
\resizebox{0.99\textwidth}{!}{
\begin{tabular}{@{}ll|ll@{}}
\toprule
Notation & Description & Notation & Description  \\ \midrule

$n$ & Dimensionality of a benchmark. & $G$ & The solution generator. \\
fitness & Goodness of a solution, related to the value of $f$. & $D$ & Bi-discriminators. \\
$U(\Omega)$ & Uniform distribution on domain $\Omega$. & $D_I$ & Exploitation discriminator. \\
$\boldsymbol{x}_\mathrm{uni}$ & Solution set sampled from $U(\Omega)$. & $D_R$ & Exploration discriminator. \\
$\boldsymbol{x}_\mathrm{samp}$ & Solution set sampled from $\boldsymbol{x}_\mathrm{opt}$ by bootstrap method. & $\boldsymbol{\omega_G}$, $\boldsymbol{\omega_I}$, $\boldsymbol{\omega_R}$ & Network parameter of $G$ , $D_{I}$, $D_{R}$. \\
$\boldsymbol{x}_\mathrm{opt}$ & Historical best solution set. & $PreIter$ & Number of generator pre-training iterations. \\
$\boldsymbol{x}_\mathrm{G}$ & Solution set generated by $G$. & $GANIter$ & Number of training iterations of GAN. \\
$M$ & Population size of $\boldsymbol{x}_\mathrm{G}$ when updating. & $DIter$ & Number of training iterations of $D$. \\
$K$ & Initial size of $\boldsymbol{x}_\mathrm{opt}$. & $\beta$ & Gradient penalty factor. \\
$a$ & Shrinking rate. & $\lambda$ & Adjustment factor for exploration-exploitation trade-off. \\
$FEs$ & Number of function evaluations. & $S$ & Batch size when training GAN. \\
$MAXFes$ & Maximum of $FEs$. & $\boldsymbol{\hat{x_I}}$ & Samples uniformly sampled along straight lines between   $\boldsymbol{x}_\mathrm{G}$ and $\boldsymbol{x}_\mathrm{samp}$. \\
$\boldsymbol{\eta}$ & Random noise. & $\boldsymbol{\hat{x_R}}$ & Samples uniformly sampled along straight lines between   $\boldsymbol{x}_\mathrm{G}$ and    $\boldsymbol{x}_\mathrm{uni}$. \\
\bottomrule
\end{tabular}}
\label{tab:notations}%
\end{table*}

%% file: alg.tex
\begin{algorithm}[t]
  \caption{OPT-GAN (see Appendix Algorithm 2 for the full version)} 
  \label{alg-OPTGAN}
\begin{algorithmic}
{
  \STATE {\bfseries Input:} The size of optimal set $K$; the batch size $S$; the population size for updating optimal set $M$; the maximum number of $FEs$ $MAXFes$; the searching domain $\Omega$; the number of iterations $GANIter$ and $DIter$ for GAN and discriminators, respectively. 
    \STATE {\bfseries Init:}
     Initialize $\boldsymbol{x}_\mathrm{opt} = \{ (\boldsymbol{x}^{(1)},\ldots,\boldsymbol{x}^{(K)}) | \boldsymbol{x}^{(s)} \sim \emph{U}(\Omega), \, s=1,2,\cdots,K \} $; 
     $MaxEpoch = \lceil ( MAXFes - K ) / M \rceil$.
  
   \STATE \textbf{Generator Pre-training:}\\ Training solution generator $G$ and discriminator $D_R$ by Eq. \ref{eq-LossG-pretraing} and Eq. \ref{eq-LossD-Exploration}, with $\emph{U}(\Omega)$.

\STATE \textbf{Distribution Reshaping:}
 \FOR{ $epoch=1$ {\bfseries to} $MaxEpoch$}
  \FOR { $iter_{G}=1$ {\bfseries to} $GANIter$}
     \FOR { $iter_{D}=1$ {\bfseries to} $DIter$}
  \STATE Training discriminator $D_{I}$ by $\mathcal{L}_\mathrm{D_{I}}$ in Eq. \ref{eq-LossD-Exploitation}, with $ {\boldsymbol{x}_\mathrm{opt}}$.
 \STATE Training discriminator $D_{R}$ by $\mathcal{L}_\mathrm{D_{R}}$ in Eq. \ref{eq-LossD-Exploration}, with $\emph{U}(\Omega)$.
 \ENDFOR

   \STATE Training solution generator $G$ by $\mathcal{L}_\mathrm{G}$ in Eq. \ref{eq-LossG}.
    
  \ENDFOR
   \STATE \textbf{Updating:}
    \STATE $\boldsymbol{x}_\mathrm{G} = G(\boldsymbol{\eta} )$; 
   $\boldsymbol{B} = \boldsymbol{x}_\mathrm{opt}  \cup \boldsymbol{x}_\mathrm{G}$. 
    \STATE Sort $\boldsymbol{B}$ according to $f(\boldsymbol{B})$ in ascending pattern.
    \STATE $\boldsymbol{x}_\mathrm{opt} = \boldsymbol{B}^{(1:K)}$.

   \STATE \textbf{Shrinking} $\boldsymbol{x}_\mathrm{opt}$ with $K$ in Eq. \ref{eq-adjust-capacity}.
   \ENDFOR
   \STATE {\bfseries Return:} ${\mathop{\arg \min }} \, f(\boldsymbol{x}_\mathrm{opt})$
}
\end{algorithmic}
\end{algorithm}

%% file: fig_tex/fig_distribution_process.tex
\begin{figure*}[tb]
\centering
\includegraphics[width=1.0\textwidth]{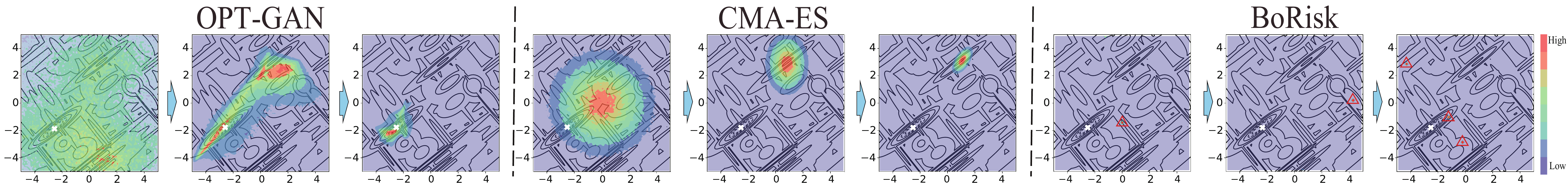}
\caption{Comparison of visualized $p(\boldsymbol{x})$ during optimization between OPT-GAN, CMA-ES, and BoRisk (see Appendix E for the plotting details of $p(\boldsymbol{x})$). The white $\times$ denotes the $x^*$. The red $\triangle$ marks the peaks of $p(\boldsymbol{x})$ for the best view.}
\label{fig:DistributionProcess_all}
\end{figure*}

%% file: fig_tex/fig_trandional_compare.tex
\begin{figure*}[tb]
    \centering
    \setlength{\belowcaptionskip}{-8 pt}
    \includegraphics[width=1.0\textwidth]{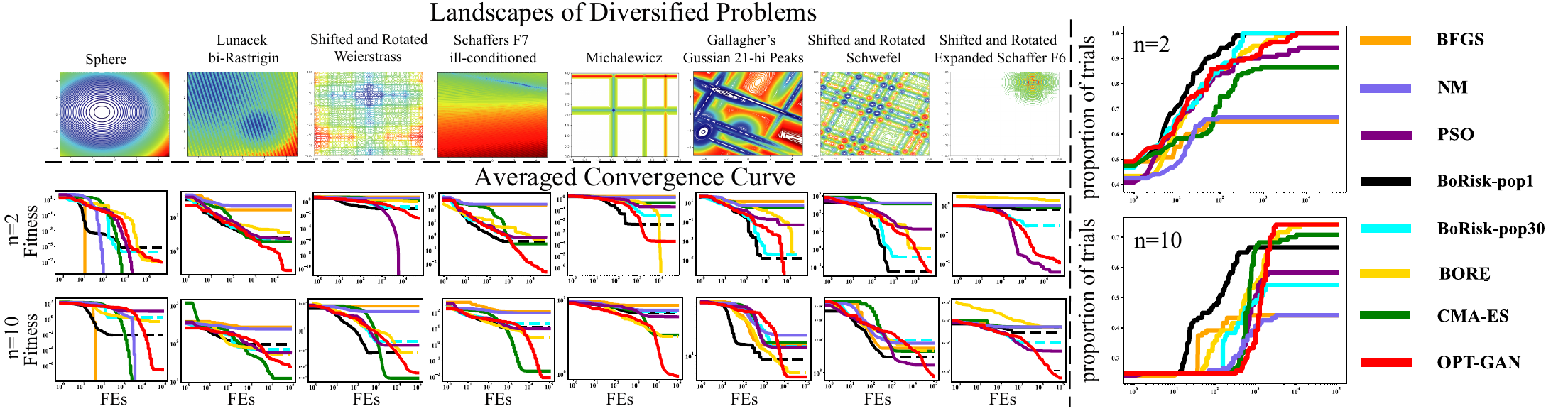}
    \caption{Average convergence curves and ECDF results of OPT-GAN and traditional optimizers on 8 benchmarks in 2 and 10 dimensions. The dotted line indicates that the optimizer stops because its elapsed time exceeds 3 hours.
    The PT for an optimizer equals the number of trials whose fitness reaches the precision  divided by the total number of trials (defined in Appendix D.1).}
    \label{fig:converg-trandional}
\end{figure*}

%% file: fig_tex/fig_nn_dim40_ecdf_hisg.tex
\begin{figure*}[htb]
  \centering
    \setlength{\belowcaptionskip}{-4pt}
	\includegraphics[width=1.0 \textwidth]{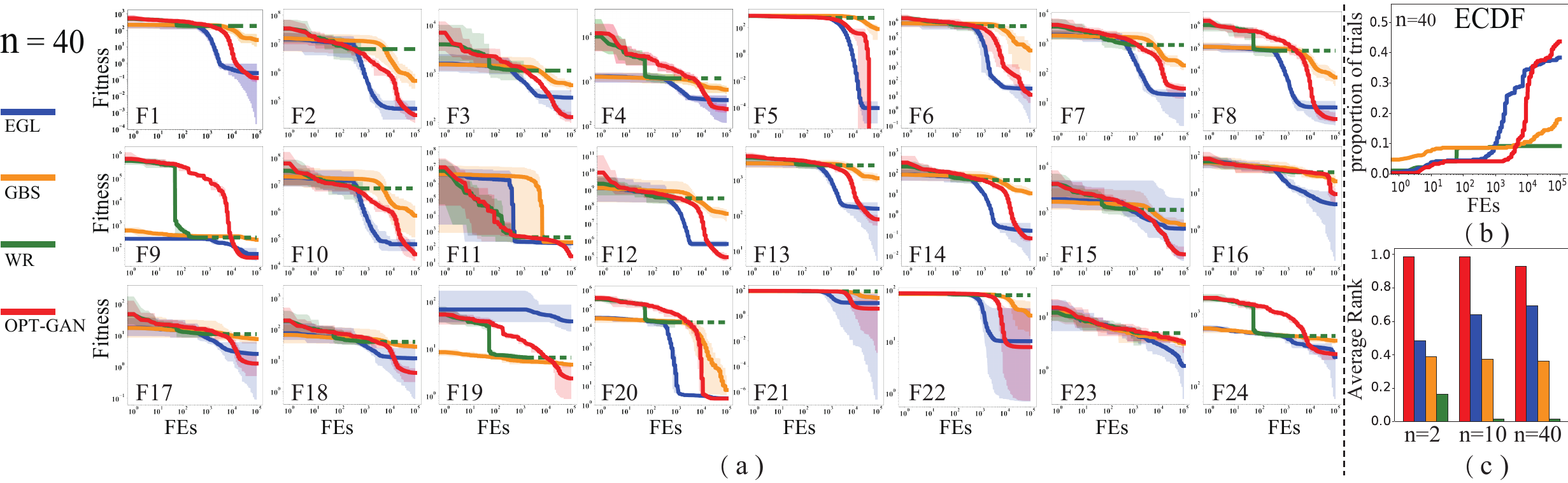}
	\caption{Performance results about (a) average convergence curves, (b) ECDF, and (c) average ranks of compared optimizers on the COCO test problems. The dotted line indicates that the optimizer stops due to elapsed time exceeding 3 hours. The shaded area reflects the ``Range". The results on benchmarks with n=2 and n=10 are shown in Appendix D.6.}
	\label{fig:converg-nn-dim40-ecdf-hisg}
\end{figure*}

%% file: fig_tex/fig_statistic_ave_rank.tex
\begin{figure}[tb]
    \centering
    \setlength{\belowcaptionskip}{-12 pt}
	\includegraphics[width=0.9\columnwidth]{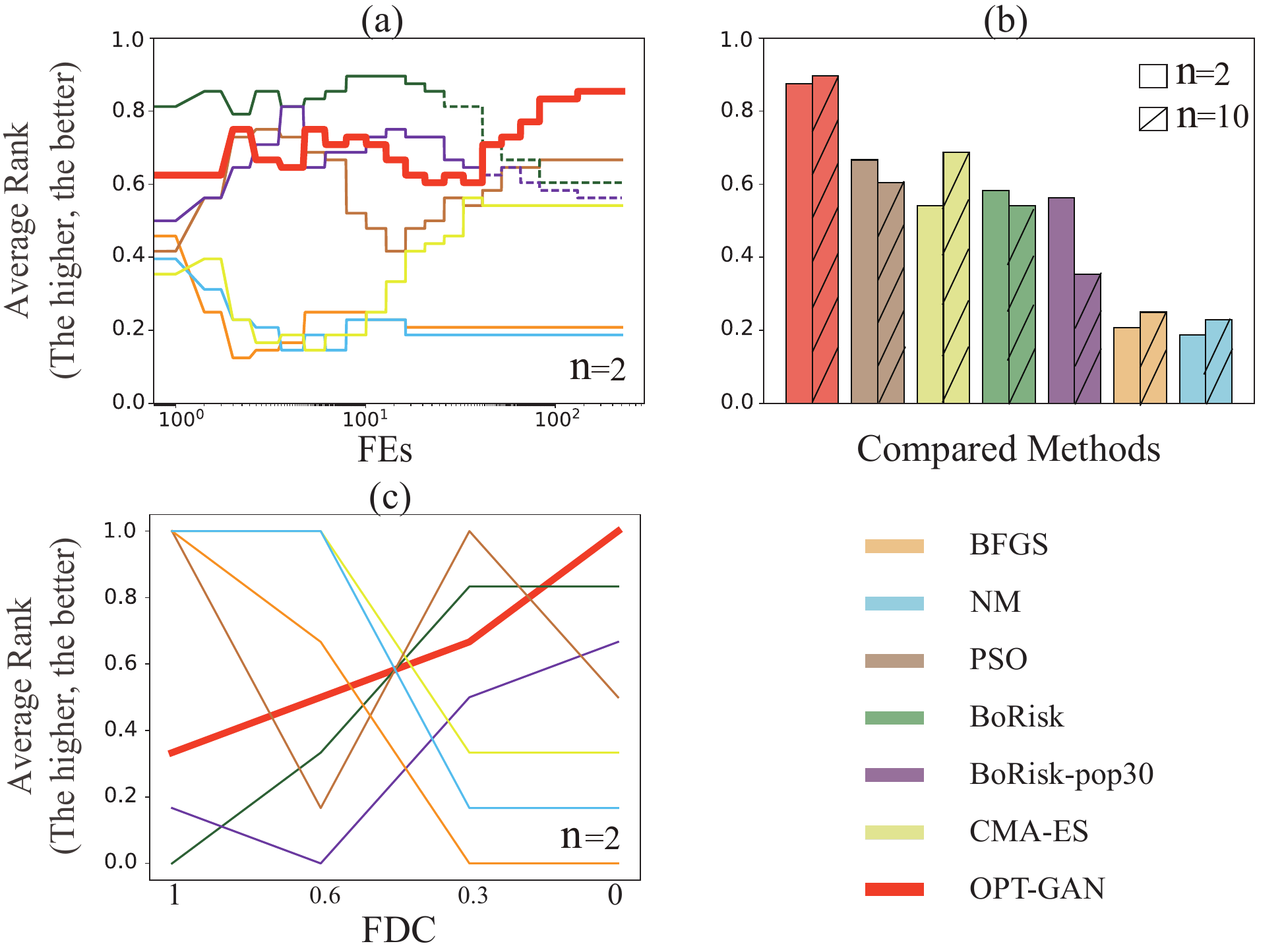}
	\caption{Average rank results versus FEs, dimensionality, and FDC.
	 The ranks of optimizers are normalized into [0, 1], where the best rank of optimizers is 1; the worst is 0.}
	\label{fig:fig-ave_rank_statistic}
\end{figure}

%% file: fig_tex/fig_realword.tex
\begin{figure*}[htb]
    \centering
    \includegraphics[width=1.0 \textwidth]{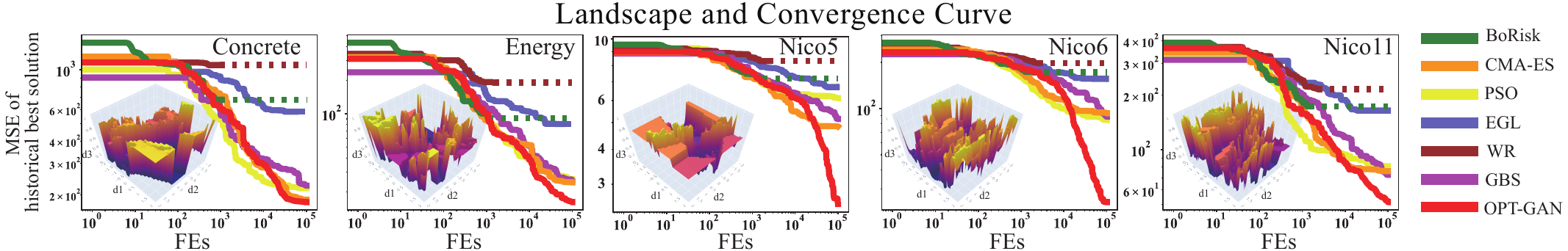}
    \caption{{The 2D projected landscapes\cite{li2018visualizing} of optimizing NEEP and the average convergence curves of optimizers.} Given the huge difference in mean square error (MSE) between different solutions in symbolic regression, the normalized rank of MSE (normalized into [0, 1], where the lowest MSE is 0 and the highest is 1) is used as the d3-axis to best view the landscapes. The dotted line indicates that the optimizer stops due to elapsed time exceeding 10 hours.}
    \label{fig:converg-realworld}
\end{figure*}

%% file: fig_tex/fig_FM.tex
\begin{figure}[tb]
    \centering
    \setlength{\belowcaptionskip}{-12 pt}
	\includegraphics[width=0.95\columnwidth]{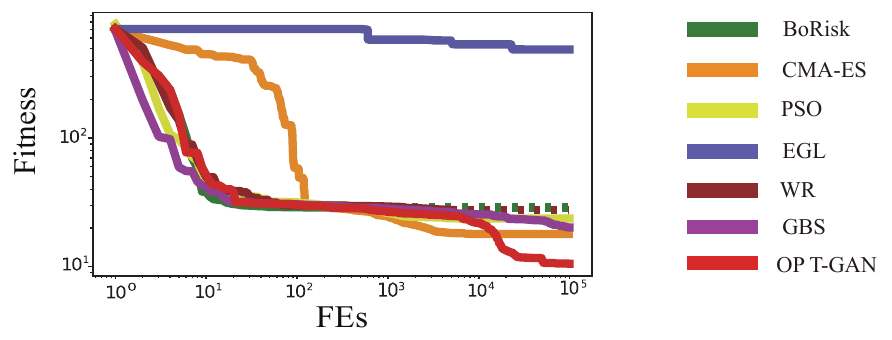}
	\caption{The average convergence curves of different optimizers in frequency modulation sound parameter identification problem. The dotted line indicates that the optimizer stops due to elapsed time exceeding 10 hours.}
	\label{fig:fig_FM}
\end{figure}

%% file: fig_tex/fig_AllDistance.tex
\begin{figure}[tb]
    \setlength{\belowcaptionskip}{-12 pt}
    \centering
	\includegraphics[width=0.95\columnwidth]{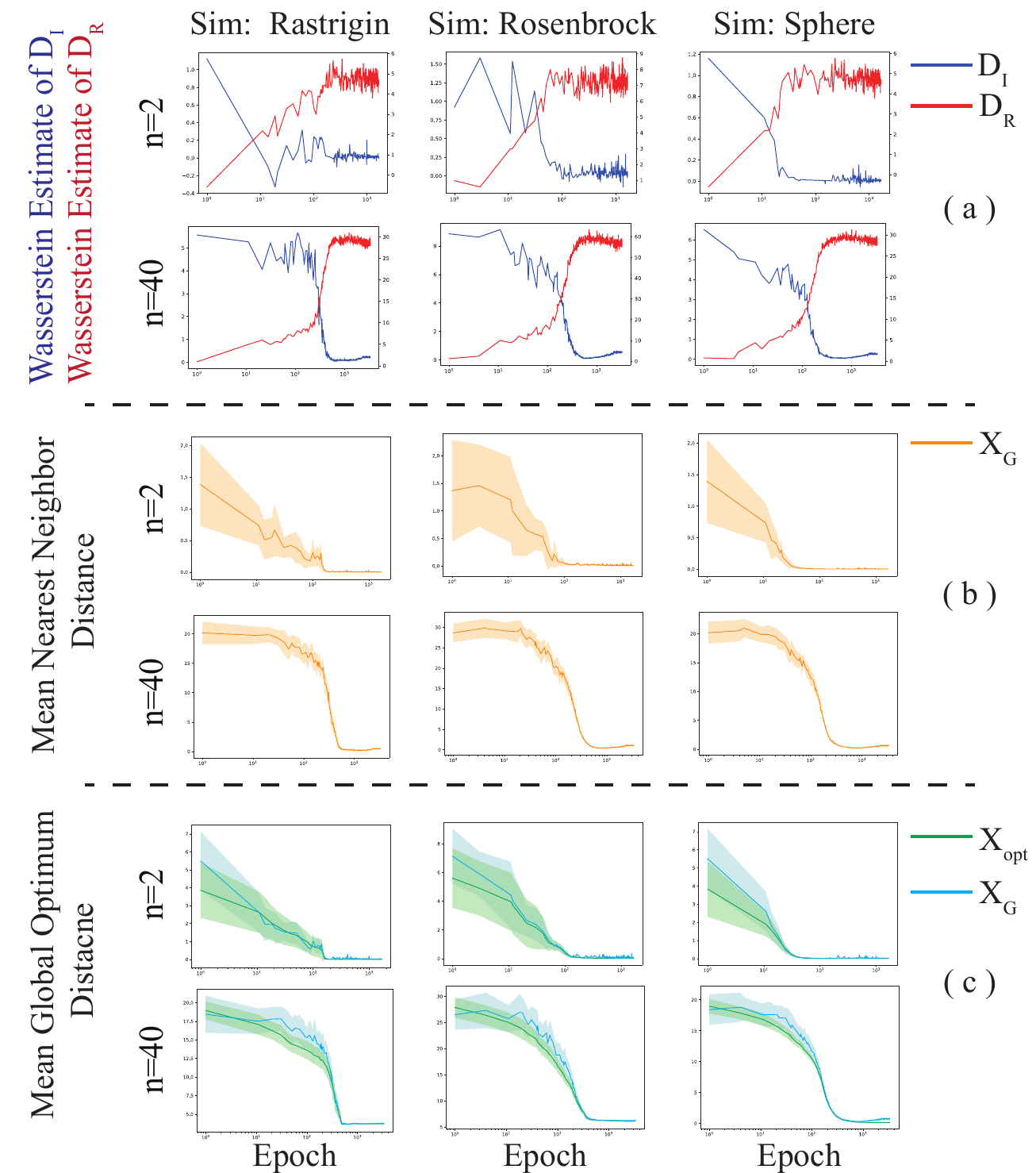}
	\caption{The convergence characteristics of OPT-GAN.
 The shaded area of curves reflects the standard deviation.}
	\label{fig:fig_AllDistance}
\end{figure}

%% file: fig_tex/fig_para.tex
\begin{figure}[t]
    \centering
    \setlength{\belowcaptionskip}{-11 pt}
    \includegraphics[width=0.95\columnwidth]{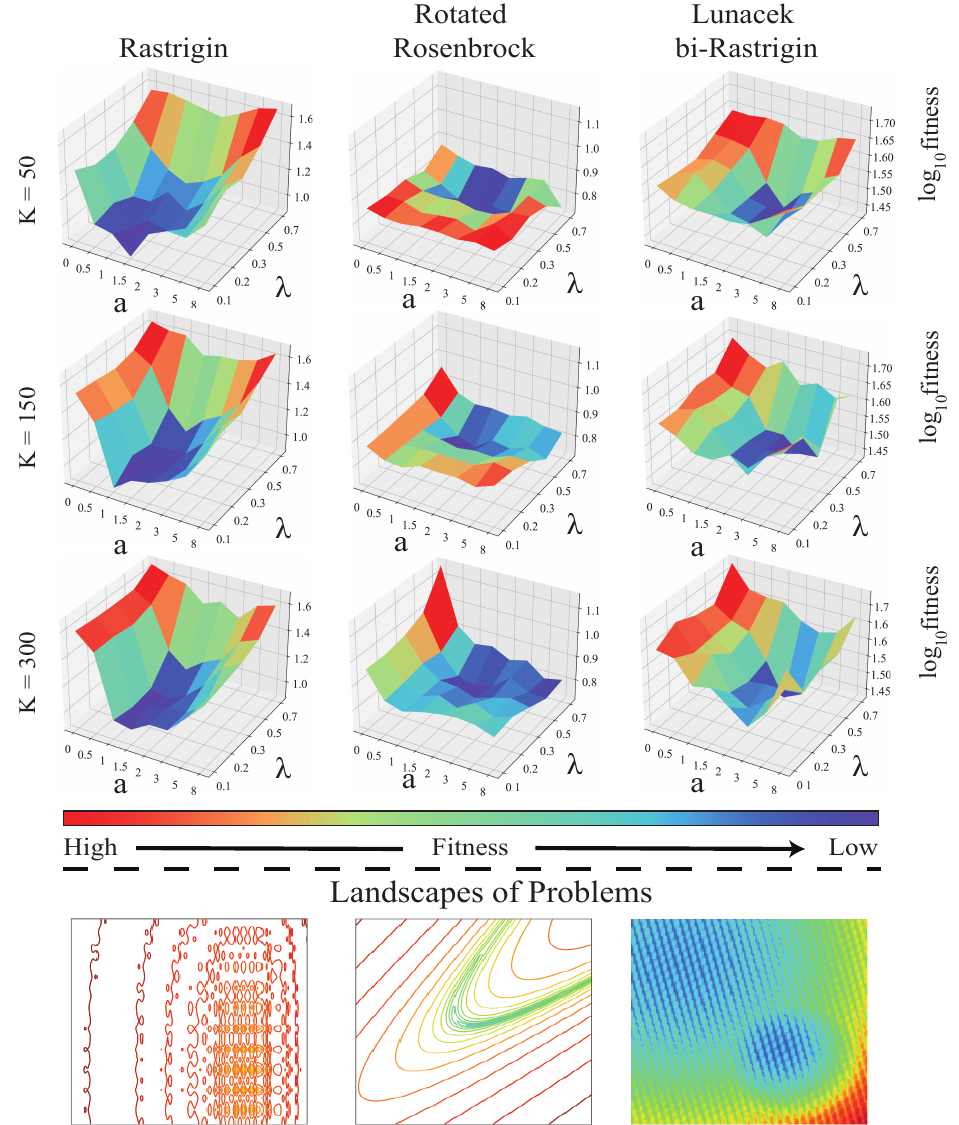}
    \caption{Hyperparameter analysis of $\boldsymbol{x}_\mathrm{opt}$'s initial size $K$, shrinking rate $a$, and adjustment factor $\lambda$. 
    }
    \label{fig:converg-para}
\end{figure}

%% file: supplemental/table_tex/opt_process.tex
\begin{algorithm}[ht]
\caption{General Optimization Framework}
  \label{General_optimization_framework}
\begin{algorithmic}
\scriptsize{
   \STATE 
   Initialize historical information base $\mathcal{D}_0$.

    \FOR{ epoch $t$= 0, 1, 2, ...}
        \STATE 1) Sampling new independent solutions from the $p(\boldsymbol{x}|\mathcal{D}_t)$;
        \STATE 2) Evaluating the new sampling solutions on $f$;
        \STATE 3) Updating the historical information base $\mathcal{D}_{t+1}$;
        \STATE 4) Updating the distribution $p(\boldsymbol{x}|\mathcal{D}_{t+1})$;
  	\ENDFOR

}
\end{algorithmic}
\end{algorithm}

%% file: supplemental/alg.tex
\begin{algorithm}[p]
  \caption{OPT-GAN}
  \label{alg-OPTGAN}
\begin{algorithmic}  
\scriptsize{
  \STATE {\bfseries Input:} The size of optimal set is $K$, the adjustment factor $\lambda$, the shrinking rate $a$, the batch size is $S$, the population size for updating optimal set is $M$, the maximum number of fitness evaluations is $MAXFes$, the number of training iterations to GAN and $D=\{D_I, D_R\}$ are $GANIter$ and $DIter$ respectively, the number of iterations of generator pre-training $PreIter$.
   \STATE {\bfseries Init:}
   \STATE Initialize $\boldsymbol{\omega_I}$, $\boldsymbol{\omega_R}$  for $D_{I}$, $D_{R}$, and $\boldsymbol{\omega_G}$ for $G$,  respectively.
   \STATE $\boldsymbol{x}_\mathrm{opt} = \{ (\boldsymbol{x}^{(1)},\ldots,\boldsymbol{x}^{(K)}) | \boldsymbol{x}^{(s)} \sim U(\Omega), \, s=1,2,\cdots,K \} $; $MaxEpoch=\lceil ( MAXFes - K ) / M \rceil$
   \STATE $t=K$; $K^{(0)}=K$
   \STATE \textbf{Generator Pre-training:}
\FOR{$iterP=1$ {\bfseries to} $PreIter$}
    \FOR{$iterG=1$ {\bfseries to} $GANIter$}
        \FOR{$iterD=1$ {\bfseries to} $DIter$}
        \STATE $\boldsymbol{x}_\mathrm{uni} = \{ (\boldsymbol{x}^{(1)}, \ldots,\boldsymbol{x}^{(S)}) \  | \  \boldsymbol{x}^{(s)} \sim U(\Omega), \, s=1,2,\cdots,S \}$
        \STATE $\boldsymbol{x}_\mathrm{G} = G_{\boldsymbol{\omega_G}}(\boldsymbol{\eta})$
        
        \STATE $\boldsymbol{\hat{x_R}}= \boldsymbol{\xi} \cdot \boldsymbol{x}_\mathrm{G} +(1-\boldsymbol{\xi} ) \cdot \boldsymbol{x}_\mathrm{uni}, \, \boldsymbol{\xi} = \{ (\boldsymbol{\xi}^{(1)}, \ldots,\boldsymbol{\xi}^{(S)}) \, | \, \boldsymbol{\xi}^{(s)} \sim U(\mathbf{0},\mathbf{1}), \, s=1,2,\cdots,S \}$
        \STATE $\boldsymbol{\omega_R} = \underset{\boldsymbol{\omega_R}}{\mathop{\arg \min }} [\frac{1}{S}\sum\limits_{s=1}^S{( {D_{R}}_{\omega_R}(\boldsymbol{x}_\mathrm{G}^{(s)}) - {D_{R}}_{\omega_R}(\boldsymbol{x}_\mathrm{uni}^{(s)})} {+ \beta ( \left\| \nabla_\mathrm{\boldsymbol{\hat{x_R}}} {D_{R}}_{\omega_R}(\boldsymbol{\hat{x_R}}^{(s)}) \right\|_2 - 1)^2})]$
        
        \ENDFOR
        \STATE $\boldsymbol{x}_\mathrm{G} = G_{\boldsymbol{\omega_G}}(\boldsymbol{\eta}) $
        \STATE $ \boldsymbol{\omega_G} =  \underset{\boldsymbol{\omega_G}}{\mathop{\arg \min }}[ -\frac{1}{S}\sum\limits_{s=1}^{S}D_{R}(\boldsymbol{x}_\mathrm{G}^{(s)})]$
    \ENDFOR
\ENDFOR
\STATE \textbf{Distribution Reshaping:}
   \FOR{ $epoch=1$ {\bfseries to} $MaxEpoch$}
  \FOR { $iterG=1$ {\bfseries to} $GANIter$}
  \STATE \textbf{Training Discriminators $D_{I}$ and $D_{R}$:}
  \FOR { $iterD=1$ {\bfseries to} $DIter$}
  \STATE  $ {\boldsymbol{x}_\mathrm{samp}} = {\{\boldsymbol{x}^{(1)}, \ldots, \boldsymbol{x}^{(S)}\}} $, sampled from $ {\boldsymbol{x}_\mathrm{opt}}$ by  bootstrapping method.
\STATE $\boldsymbol{x}_\mathrm{uni} = \{ (\boldsymbol{x}^{(1)}, \ldots,\boldsymbol{x}^{(S)}) | \boldsymbol{x}^{(s)} \sim U(\Omega), \, s=1,2,\cdots,S \}$
  \STATE $\boldsymbol{x}_\mathrm{G} = G_{\boldsymbol{\omega_G}}(\boldsymbol{\eta})$
    \STATE $\boldsymbol{\hat{x_I}}= \boldsymbol{\xi} \cdot \boldsymbol{x}_\mathrm{G} +(1-\boldsymbol{\xi} ) \cdot \boldsymbol{x}_\mathrm{samp}, \, \boldsymbol{\xi} = \{ (\boldsymbol{\xi}^{(1)}, \ldots,\boldsymbol{\xi}^{(S)}) \, | \, \boldsymbol{\xi}^{(s)} \sim U(\mathbf{0},\mathbf{1}), \, s=1,2,\cdots,S \}$
    \STATE $\boldsymbol{\hat{x_R}}= \boldsymbol{\xi} \cdot \boldsymbol{x}_\mathrm{G} +(1-\boldsymbol{\xi} ) \cdot \boldsymbol{x}_\mathrm{uni}, \, \boldsymbol{\xi} = \{ (\boldsymbol{\xi}^{(1)}, \ldots,\boldsymbol{\xi}^{(S)}) \, | \, \boldsymbol{\xi}^{(s)} \sim U(\mathbf{0},\mathbf{1}), \, s=1,2,\cdots,S \}$
    \STATE $\boldsymbol{\omega_I} = \underset{\boldsymbol{\omega_I}}{\mathop{\arg \min }} [\frac{1}{S}\sum\limits_{s=1}^S{( {D_{I}}_{\omega_I}(\boldsymbol{x}_\mathrm{G}^{(s)}) - {D_{I}}_{\omega_I}(\boldsymbol{x}_\mathrm{samp}^{(s)})} {+ \beta ( \left\| \nabla_\mathrm{\boldsymbol{\hat{x_I}}} {D_{I}}_{\omega_I}(\boldsymbol{\hat{x_I}}^{(s)}) \right\|_2 - 1)^2})]$
    \STATE $\boldsymbol{\omega_R} = \underset{\boldsymbol{\omega_R}}{\mathop{\arg \min }} [\frac{1}{S}\sum\limits_{s=1}^S{( {D_{R}}_{\omega_R}(\boldsymbol{x}_\mathrm{G}^{(s)}) - {D_{R}}_{\omega_R}(\boldsymbol{x}_\mathrm{uni}^{(s)})} {+ \beta ( \left\| \nabla_\mathrm{\boldsymbol{\hat{x_R}}} {D_{R}}_{\omega_R}(\boldsymbol{\hat{x_R}}^{(s)}) \right\|_2 - 1)^2})]$
  
  \ENDFOR
   \STATE \textbf{Training Solution Generator $G$:}
  \STATE $\boldsymbol{x}_\mathrm{G} = G_{\boldsymbol{\omega_G}}(\boldsymbol{\eta} )$
  \STATE $ \boldsymbol{\omega_G} =  \underset{\boldsymbol{\omega_G}}{\mathop{\arg \min }}[ -\frac{1}{S}\sum\limits_{s=1}^{S}(\frac{1}{1+ \lambda}{D_{I}}_{\omega_I}(\boldsymbol{x}_\mathrm{G}^{(s)}) + \frac{\lambda}{1+ \lambda} {D_{R}}_{\omega_R}(\boldsymbol{x}_\mathrm{G}^{(s)}))]$
  
  \ENDFOR
   \STATE \textbf{Updating:}
   \STATE $\boldsymbol{x}_\mathrm{G} = G_{\boldsymbol{\omega_G}}(\boldsymbol{\eta})$    
   {
   \STATE $\boldsymbol{B} = \boldsymbol{x}_\mathrm{opt}  \cup \boldsymbol{x}_\mathrm{G}$}.
   \STATE Sort $\boldsymbol{B}$ according to their $f(\boldsymbol{x}_\mathrm{B})$ in ascending pattern.
   \STATE $t = t+M$
   \STATE \textbf{Shrinking:} 
   \STATE $K^{(t)} = \lceil {{K}^{(0)}}^{(1- a \cdot (t/MAXFes))} \rceil$
   {
   \STATE $\boldsymbol{x}_\mathrm{opt} = \boldsymbol{B}^{(1:K^{(t)})}$
   }
   \ENDFOR

   \STATE {\bfseries Return:} ${\mathop{\arg \min }} \, f(\boldsymbol{x}_\mathrm{opt})$
}

\end{algorithmic}
\end{algorithm}

%% file: supplemental/table_tex/tab_notations.tex
\begin{table}[b]
\centering
\caption{Description of notations.}
\resizebox{1.0\columnwidth}{!}{
\begin{tabular}{@{}ll|ll@{}}
\toprule
Notation & Description & Notation & Description  \\ \midrule

$n$ & The dimensionality of a benchmark. & $G$ & The solution generator. \\
fitness & The value of $f$. & $D$ & The bi-discriminators. \\
$U(\Omega)$ & The uniform distribution on domain $\Omega$. & $D_I$ & The exploitation discriminator. \\
$\boldsymbol{x}_\mathrm{uni}$ & The solution set sampled from $U(\Omega)$. & $D_R$ & The exploration discriminator. \\
$\boldsymbol{x}_\mathrm{samp}$ & The solution set sampled from $\boldsymbol{x}_\mathrm{opt}$  by bootstrap method. & $\boldsymbol{\omega_G}$, $\boldsymbol{\omega_I}$, $\boldsymbol{\omega_R}$ & The network parameter of $G$ , $D_{I}$, $D_{R}$. \\
$\boldsymbol{x}_\mathrm{opt}$ & The historical best solution set. & $PreIter$ & The number of generator pre-training iterations. \\
$\boldsymbol{x}_\mathrm{G}$ & The solution set generated by $G$. & $GANIter$ & The number of training iterations of GAN. \\
$M$ & The population size of $\boldsymbol{x}_\mathrm{G}$ when updating. & $DIter$ & The number of training iterations of $D$. \\
$K$ & The size of $\boldsymbol{x}_\mathrm{opt}$. & $\beta$ & The gradient penalty factor. \\
$a$ & The shrinking rate. & $\lambda$ & The adjustment factor for exploration-exploitation trade-off. \\
$FEs$ & The number of function evaluations. & $S$ & The batch size when training GAN. \\
$MAXFes$ & The maximum of $FEs$. & $\boldsymbol{\hat{x_I}}$ & It is sampled uniformly along straight lines between   $\boldsymbol{x}_\mathrm{G}$ and $\boldsymbol{x}_\mathrm{samp}$. \\
$\boldsymbol{\eta}$ & The random noise. & $\boldsymbol{\hat{x_R}}$ & It is sampled uniformly along straight lines between   $\boldsymbol{x}_\mathrm{G}$ and    $\boldsymbol{x}_\mathrm{uni}$. \\
\bottomrule
\end{tabular}}
\label{tab:notations}%
\end{table}

%% file: supplemental/fig_tex/fig_K.tex
\begin{figure}[tb]
    \centering
    \includegraphics[width=0.5\columnwidth]{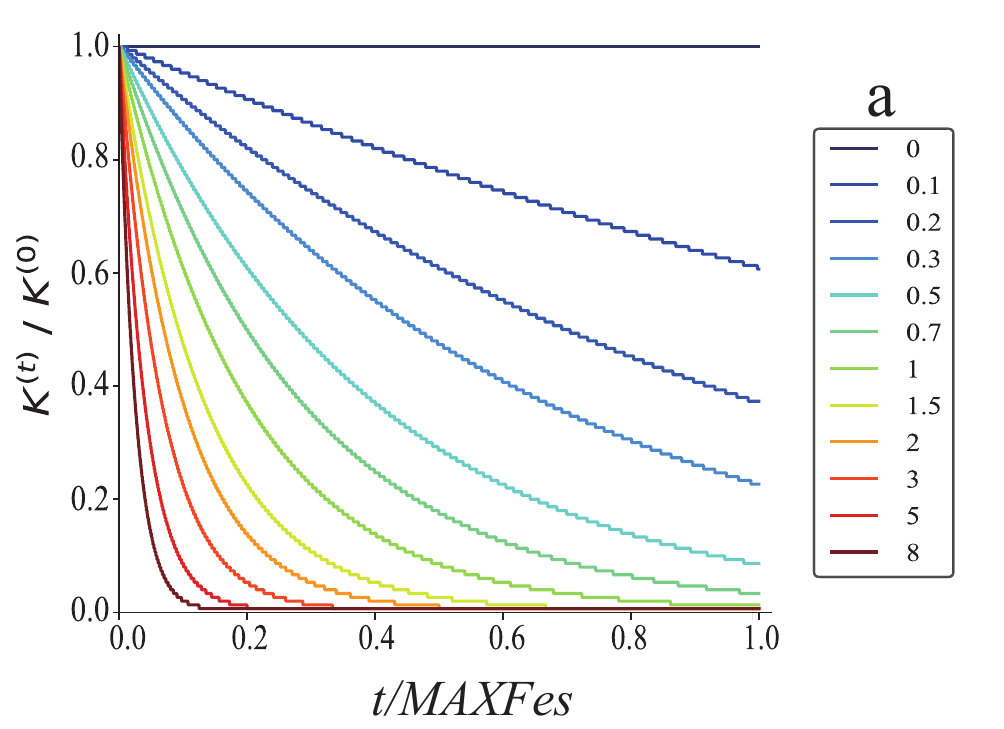}
    \caption{{Shrinking process of $K$ based on different $a$.}}
    \label{fig:k_curve}
\end{figure}

%% file: supplemental/table_tex/benchmarks_inf.tex
\begin{table}[htb]
 \caption{The information of benchmarks}
    \label{benchmarkinfo}
    \resizebox{\columnwidth}{!}{
\begin{tabular}{llc}
\toprule
     & Benchmarks Name & Search Domain $\Omega$        \\ \midrule
\begin{tabular}[c]{@{}c@{}}Platform for Comparing \\ Continuous Optimizers\\ (COCO)\end{tabular} & \begin{tabular}[c]{@{}l@{}}Sphere (F1), Ellipsoidal (F2), Rastrigin (F3),\\ Buche-Rastrigin (F4), Linear Slope (F5), \\ Attractive Sector (F6), Step Ellipsoidal (F7), \\ Original Rosenbrock (F8), Rotated Rosenbrock (F9),\\ non-separable Ellipsoida (F10), Discus (F11), \\ Bent Cigar (F12), Sharp Ridge (F13), \\Different Powers (F14),  non-separable Rastrigin (F15), \\Weierstrass (F16),  Schaffers F7 (F17),\\ Schaffers F7 ill-conditioned (F18), \\ Griewank-Rosenbrock (F19), Schwefel (F20), \\ Gallagher's Gaussian 101-me Peaks (F21),\\ Gallagher's Gaussian 21-hi Peaks (F22), Katsuura (F23), \\ Lunacek bi-Rastrigin (F24).\end{tabular}  & $[-5,5]^n$    \\ \midrule
CEC'19 Benchmark Suite & \begin{tabular}[c]{@{}l@{}}Shifted and Rotated Schwefel, \\ Shifted and Rotated Expanded Schaffer F6, \\ Shifted and Rotated Weierstrass.\end{tabular}   & $[-100,100]^n$ \\ \midrule
Simulationlib & \begin{tabular}[c]{@{}l@{}}Michalewicz,\\ {Sim: Sphere, Sim: Rastrigin, Sim: Rosenbrock. }\end{tabular}   & \begin{tabular}[c]{@{}c@{}}$

[0,4]^n$ \\ { $[-5.12,5.12]^n$ } \end{tabular}   \\ \midrule
Conformal Bent Cigar & Conformal Bent Cigar. & $[-5,5]^n$ \\ \bottomrule
\end{tabular}
}
\end{table}

%% file: supplemental/table_tex/tab_exp_setting.tex
\begin{table}[htb]
    \centering
    \begin{threeparttable}
    \caption{{Hyperparameter settings of compared methods}}
    \label{tab:ParameterSettings}%
    \begin{tabularx}{\columnwidth}{p{0.2\columnwidth}X}
    \toprule
    OPT-GAN & The optimal set size $K = 150$, the population size $M = 30$, the shrinking rate $a = 1.5$, the adjustment factor $\lambda = 0.3$. The number of training iterations of GAN $GANIter = 150$, the number of training iterations of discriminators $DIter = 4$, the number of training iterations in pre-training $PreIter = 100$, the gradient penalty factor $\beta = 0.1$, the batch size $S = 30$.
    Adam is used as the training method. The learning rate of $G$ is 0.0001, and the learning rate of $D=\{D_I, D_R\}$ is 0.005. \newline
    The code of OPT-GAN is implemented based on the pytorch (version 1.10.2) \cite{NEURIPS2019_9015} package, which released under the BSD license.
    \\
    \midrule
     GBS & The number of drawn samples is 100, the percentile is 0.5. 
    \newline The structure of discriminator is MLP with single hidden layer, and the structure of generator is MLP with two hidden layers. Relu and sigmoid functions are adopted into the hidden layer and output layer, respectively. The number of neurons in the hidden layer is 100. \newline
    The code of GBS is implemented based on the pytorch package.\\
    \midrule
    WR & The population size $M = 30$. \newline 
    Other hyperparameters follow the settings from the paper\cite{NEURIPS2020_81e3225c}. \newline 
    The code of WR provides from authors. \\
    \midrule
    CMA-ES, NM, \newline BFGS, EGL & The population size $M = 30$.  \newline 
    Other hyperparameters follow the settings from the paper \cite{Sarafian2020EGL}.  \newline 
    The code of CMA-ES is implemented based on the pycma package. \newline 
    The code of NM, BFGS are implemented based on the Scipy package. \newline 
    The code of EGL provides by authors. \\
    \midrule
    PSO & The population size $M = 30$. \newline 
    Other hyperparameters follow the settings from the paper\cite{tang2021novel}. \newline 
    The code is implemented based on the scikit-opt package.
    \\
    \midrule
    BoRisk & BoRisk: Hyperparameters follow the setting from the paper\cite{BoRisk}.
    \newline
    BoRisk-pop30: The size of initial set  $num\_samp = 150$, the size of candidates $candidate = 30$ at each iteration; other hyperparameters are same with  BoRisk. \newline
    Their codes provide from authors, which are implemented based on the Botorch package.
     \\
    \bottomrule
     \end{tabularx}
    
    \end{threeparttable}
\end{table}%

%% file: supplemental/fig_tex/fig_ablation.tex
\begin{figure}[tb]
    \centering
    \includegraphics[width=1.0\columnwidth]{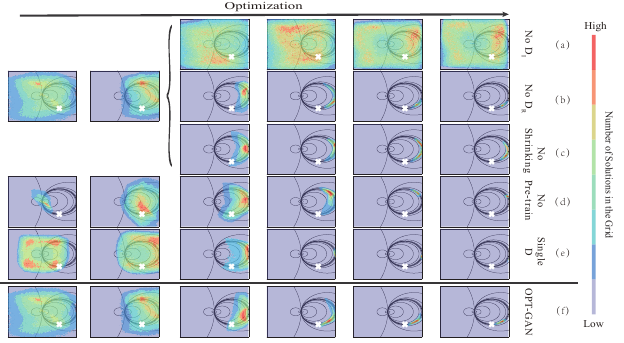}
    \caption{Ablation experiment for exploration discriminator $D_R$, exploitation discriminator $D_I$, generator pre-training, shrinking strategy, and bi-discriminators on the 2D Conformal Bent Cigar benchmark. Each subplot is a mixture of the landscape and heatmap. Subplots in the same column denote that they are generated with the same number of $FEs$.
    The heatmap displays the distribution denoted by $G$, and is approximated by Monte Carlo sampling with one million samples. The black $\times$ denotes the position of the optimum. For No-$D_R$ OPT-GAN, No-$D_I$ OPT-GAN,  and No-Shrinking OPT-GAN, the corresponding component is removed from OPT-GAN  to observe the influence on the optimization, when $FEs$ exceeds the threshold ($FEs > 500$).  The hyperparameters are set as follows: $a =0.525$, $\lambda=0.3$, $m=30$, $K=150$, $PreIter=130$, and $MAXFes=3500$. Other hyperparameter settings are the same with those in Table \ref{tab:ParameterSettings}.}
    \label{fig:ablation}
\end{figure}

%% file: supplemental/fig_tex/fig_AllDistance.tex
\begin{figure}[tb]
    \centering
	\includegraphics[width=1\columnwidth]{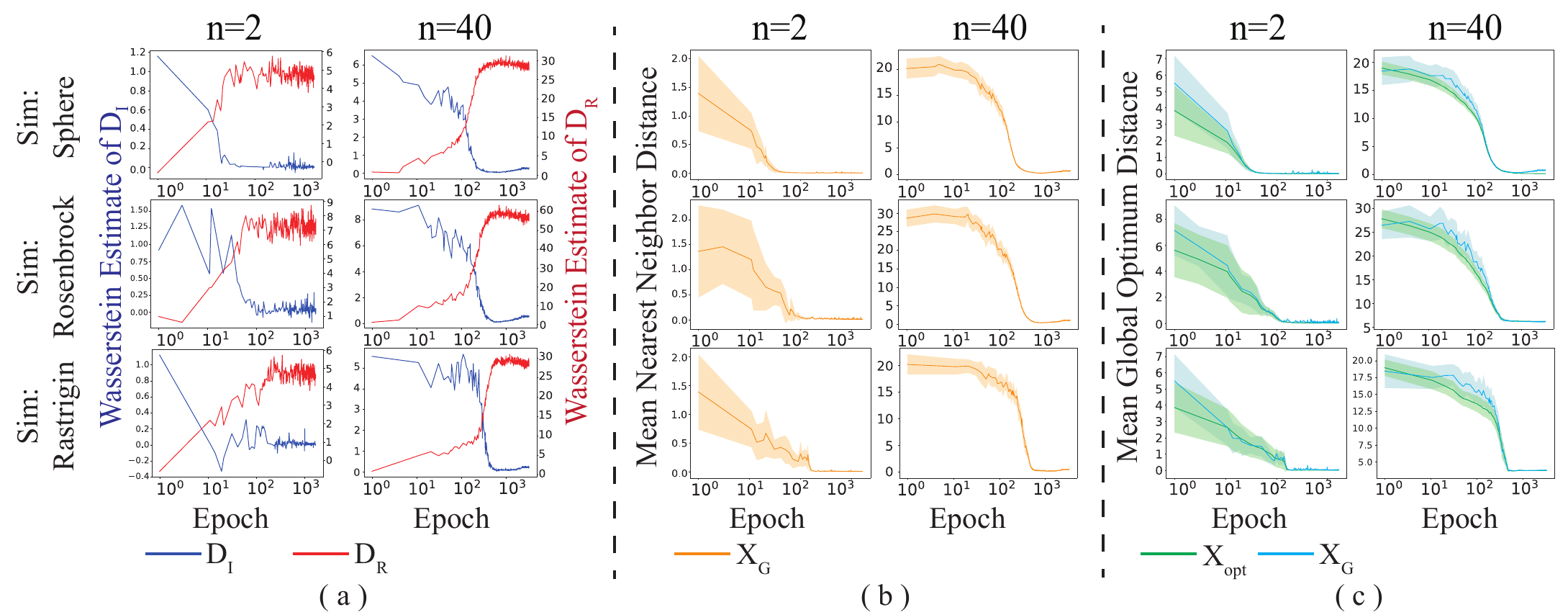}
	\caption{{Verifying the convergence of OPT-GAN through the three perspectives: (a) wasserstein estimate of $D_I$ and $D_R$; (b) the mean nearest neighbor distance curve between generated solutions $\boldsymbol{x_{G}}$; (c) the mean global optimum distance curves of the global optimum $\boldsymbol{x}^*$ to $\boldsymbol{x_{G}}$ and $\boldsymbol{x_{opt}}$ respectively. The shaded area of curves reflects the standard deviation.} }
	\label{fig:fig_AllDistance}
\end{figure}

%% file: supplemental/fig_tex/fig_para.tex
\begin{figure}[tb]
    \centering
    \includegraphics[width=0.85\columnwidth]{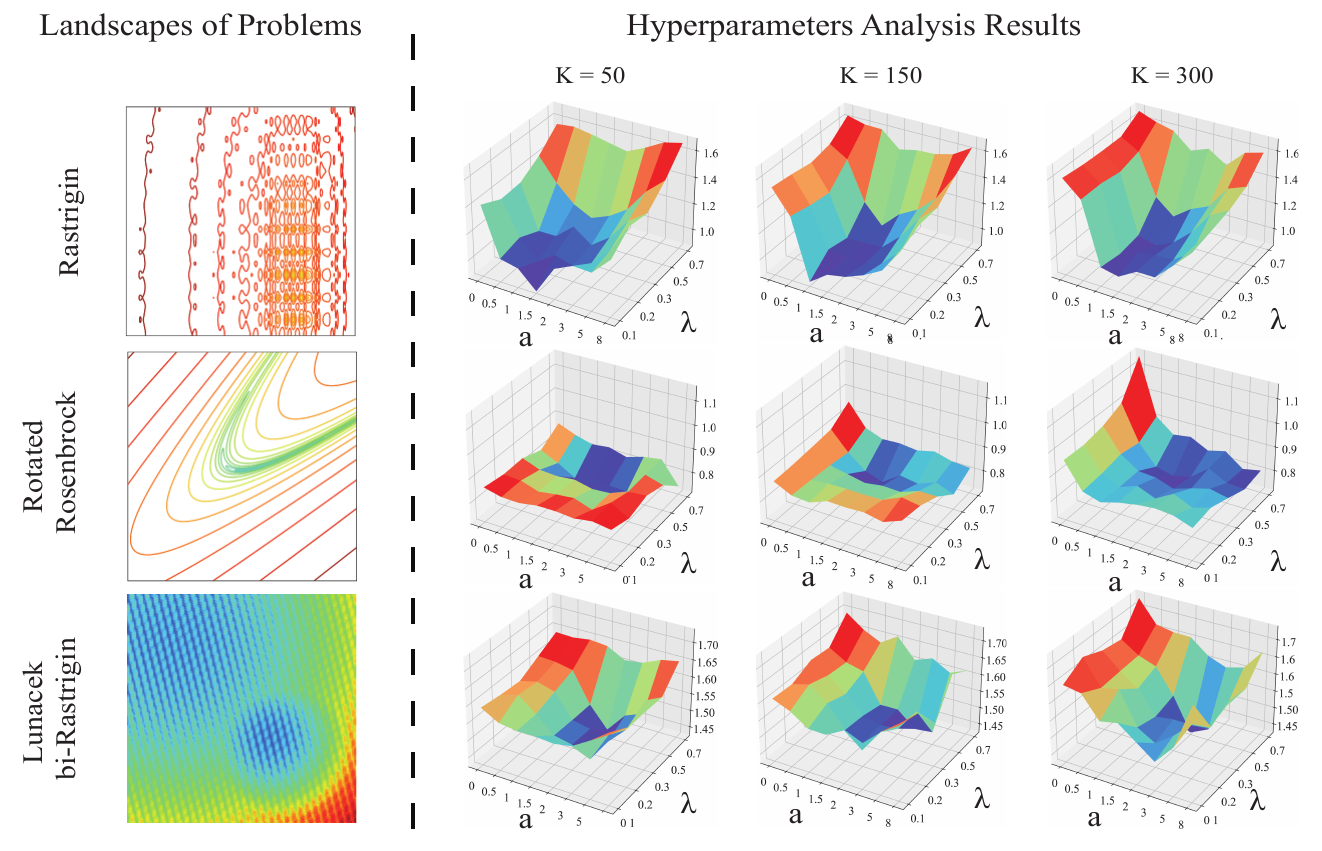}
    \caption{Hyperparameter analysis of the optimal set size $K$, shrinking rate $a$,  and adjustment factor $\lambda$. The ordinate represents the fitness, which is computed by $\log_{10}{f}$. The experimental results are derived from the average of 15 repeated trials with $MAXFes=10000$. Other hyperparameter settings are the same with those in Table \ref{tab:ParameterSettings}.}
    \label{fig:converg-para}
\end{figure}

%% file: supplemental/table_tex/tab_ttest_trandional.tex
\begin{table}[htb]
  \centering
  \caption{{Results of t-tests with the significance level of 0.05 between OPT-GAN and trandional optimizers.}}
  \resizebox{0.95\columnwidth}{!}{
    \begin{tabular}{lllllllllllllll}
    \toprule
           & \multicolumn{2}{l}{BoRisk-pop1} & \multicolumn{2}{l}{BoRisk-pop30} & \multicolumn{2}{l}{CMA-ES} & \multicolumn{2}{l}{BFGS} & \multicolumn{2}{l}{Nelder-Mead} & \multicolumn{2}{l}{PSO} & \multicolumn{2}{l}{BORE} \\
       & n=2       & n=10       & n=2        & n=10        & n=2         & n=10         & n=2       & n=10       & n=2            & n=10           & n=2        & n=10           & n=2        & n=10       \\ \midrule
    Better & 7 & 7 & 6 & 8 & 5 & 4 & 7 & 7 & 7 & 7 & 3 & 8 & 4 & 8 \\ 
        Worse & 0 & 0 & 0 & 0 & 0 & 3 & 0 & 1 & 0 & 1 & 2 & 0 & 1 & 0 \\ 
        Same & 1 & 1 & 2 & 0 & 3 & 1 & 1 & 0 & 1 & 0 & 3 & 0 & 3 & 0 \\ \midrule
        Merit & 7 & 7 & 6 & 8 & 5 & 1 & 7 & 6 & 7 & 6 & 1 & 8 & 3 & 8 \\ \bottomrule
    \end{tabular}%
  }
  \label{tab:compare-ttests-trandional}%
\end{table}%

%% file: supplemental/fig_tex/fig_nn_dim2.tex
\begin{figure*}[htb]
  \centering
	\includegraphics[width=0.8\textwidth]{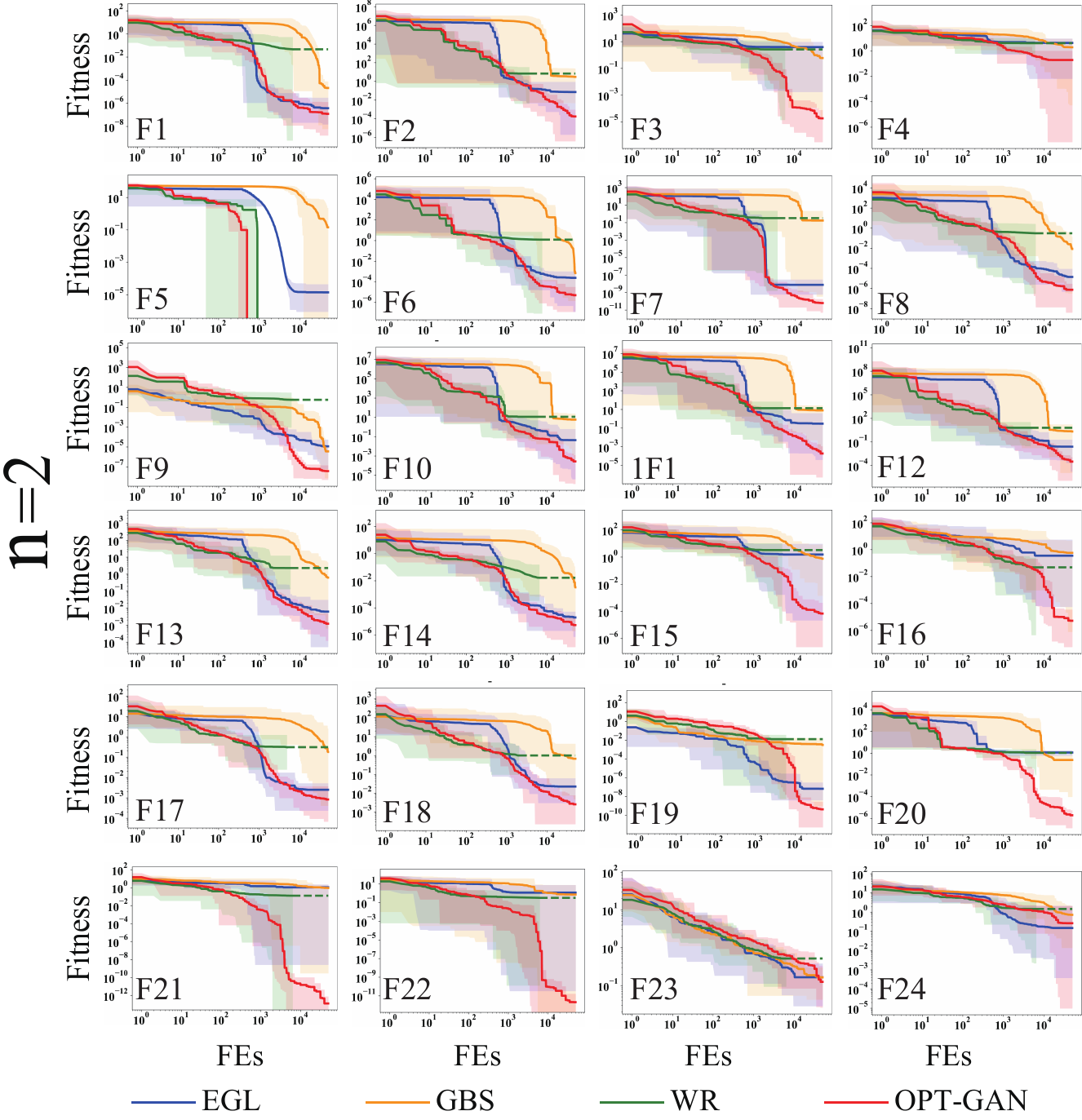}
	\caption{{Average convergence curves of OPT-GAN and NN-based optimizers on different benfchmarks in 2 dimensions.} Each curve denotes the average fitness of an optimizer on 15 repeated trials. The shadow area of each curve is limited by the worst and best solutions in each evaluation. The dotted line indicates that the optimizer stops because its elapsed time exceeds 3 hour.}
	\label{fig:converg-nn-dim2}
\end{figure*}

%% file: supplemental/fig_tex/fig_nn_dim10.tex
\begin{figure*}[htb]
  \centering
	\includegraphics[width=0.8\textwidth]{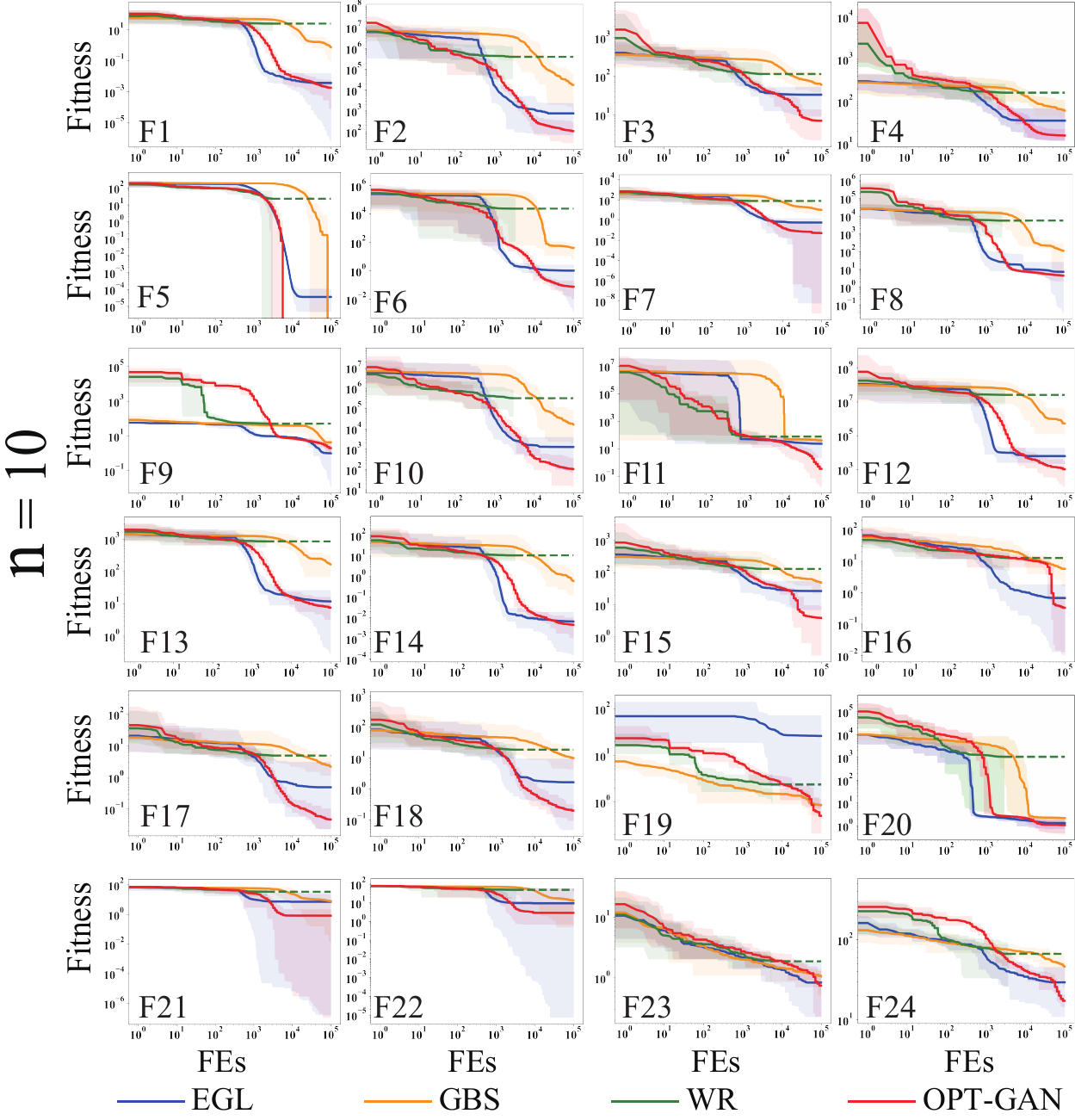}
	\caption{{Average convergence curves of OPT-GAN and NN-based optimizers on different benchmarks in 10 dimensions.} Each curve denotes the average fitness of an optimizer on 15 repeated trials. The shaded area of the averaged convergence curve reflects the ``Range".  The dotted line indicates that the optimizer stops because its elapsed time exceeds 3 hour.}
	\label{fig:converg-nn-dim10}
\end{figure*}

%% file: supplemental/fig_tex/fig_nn_ecdf+hisg.tex
\begin{figure}[tb]
    \centering
	\includegraphics[width=0.8\columnwidth]{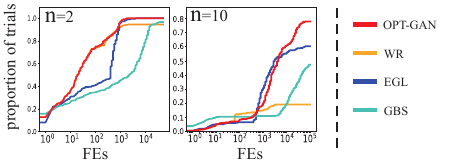}
	\caption{{The ECDF results of OPT-GAN compared with NN-based optimizers based on 15 repeated trials on 24 benchmarks in 2 and 10 dimensions.}}
	\label{fig:fig_NN_hisg}
\end{figure}

%% file: supplemental/table_tex/tab_ttest_NN.tex
\begin{table}[tb]
  \centering
  \caption{Results of t-tests  with the significance level of 0.05 between OPT-GAN and neural network-based optimizers.}
  \resizebox{0.7\columnwidth}{!}{
    \begin{tabular}{lllllllllllll}
    \hline
       & \multicolumn{3}{l}{EGL} & \multicolumn{3}{l}{GBS} &
       \multicolumn{3}{l}{WR} \\
       & n=2   & n=10   & n=40   & n=2      & n=10     & n=40     & n=2   & n=10   & n=40  \\ \midrule
Better & 16    & 15     & 16     & 8        & 23       & 23       & 19    & 23     & 24 \\
Worse  & 0     & 1      & 3      & 0        & 0        & 1        & 0     & 0      & 0    \\
Same   & 8     & 8      & 5      & 16       & 1        & 0        & 5     & 1      & 0     \\ \midrule
Merit  & 16    & 14     & 13     & 8        & 23       & 22       & 19    & 23     & 24    \\ \bottomrule
        
    \end{tabular}%
  }
  \label{tab:compare-ttests}%
\end{table}%

%% file: supplemental/table_tex/tab_time.tex
\begin{table}[tb]
    \centering
    \caption{Time consumption of OPT-GAN and learning-based optimizers.}
    \resizebox{0.8\columnwidth}{!}{
    
    \begin{threeparttable}
    
    \begin{tabular}{@{}lccccc@{}}
    \toprule
                                & OPT-GAN   & WR        & EGL       & GBS & BoRisk        \\ \midrule
    Rastrigin                   & 0.44h & 25.3h & 0.28h & 4.36E-4h        & 	$\gg$24h \\
    Attractive Sector           & 0.52h & 28.1h & 0.38h & 4.53E-4h        & 	$\gg$24h \\
    Different Powers             & 0.51h & 22.5h & 0.27h & 4.11E-4h        & 	$\gg$24h \\
    Schaffers F7                & 0.52h & 20.5h & 0.27h & 4.03E-4h        & 	$\gg$24h \\
    \begin{tabular}[c]{@{}l@{}}Gallagher’s Gaussian 101-me\end{tabular}  & 0.44h & 25.4h & 0.28h & 3.33E-4h        & 	$\gg$24h \\ \midrule
    Average                     & 0.49h & 24.36h & 0.30h & 4.08E-4h        & 	$\gg$24h \\ \bottomrule
    \end{tabular}%
    
    \begin{tablenotes}
        \footnotesize
        \item The \emph{FEs} is set to 10000, the dimensionality of benchmarks is 10.
        Other hyperparameter settings are the same with those in Table \ref{tab:ParameterSettings}. h means hours.
      \end{tablenotes}
    
    \end{threeparttable}
    }
    \label{tab:time-compare-chart}%
\end{table}%